\documentclass{article}

\usepackage{arxiv}

\usepackage[utf8]{inputenc} 
\usepackage[T1]{fontenc}    
\usepackage{hyperref}       
\usepackage{url}            
\usepackage{booktabs}       
\usepackage{amsfonts}       
\usepackage{nicefrac}       
\usepackage{microtype}      
\usepackage{cleveref}       
\usepackage{lipsum}         
\usepackage{graphicx}
\usepackage{natbib}
\usepackage{doi}
\usepackage{float}
\usepackage{threeparttable}

\title{A Hardware-Algorithm Co-Designed Framework for HDR Imaging and Dehazing in Extreme Rocket Launch Environments}

\usepackage{authblk}

\newbox{\orcid}\sbox{\orcid}{\includegraphics[scale=0.06]{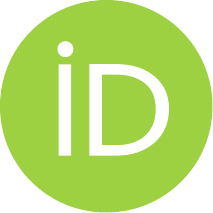}} 
\author[1,2]{Jing Tao}
\author[1,2,*]{Banglei Guan}
\author[1,2]{Pengju Sun}
\author[1,2]{Taihang Lei}
\author[1,2]{Yang Shang}
\author[1,2]{Qifeng Yu}
\affil[1]{College of Aerospace Science and Engineering, National University of Defense Technology, Hunan 410073, China}
\affil[2]{Hunan Provincial Key Laboratory of Image Measurement and Vision Navigation, Hunan 410073, China}

\renewcommand{\shorttitle}{Acta Mechanica Sinica}

\hypersetup{
	pdftitle={A Hardware-Algorithm Co-Designed Framework for HDR Imaging and Dehazing in Extreme Rocket Launch Environments},
	pdfauthor={Jing Tao, Banglei Guan, Pengju Sun, Taihang Lei, Yang Shang, Qifeng Yu},
	pdfkeywords={HDR Imaging, Dehazing, Rocket Launch, Vision Navigation, Image Measurement},
	pdfsubject={Computer Vision and Pattern Recognition},
	colorlinks=true, 
	linkcolor=red,   
	citecolor=blue,  
	urlcolor=blue    
}

\begin{document}
\maketitle

\begin{abstract}
	Quantitative optical measurement of critical mechanical parameters---such as plume flow fields, shock wave structures, and nozzle oscillations---during rocket launch faces severe challenges due to extreme imaging conditions. Intense combustion creates dense particulate haze and luminance variations exceeding 120 dB, degrading image data and undermining subsequent photogrammetric and velocimetric analyses. To address these issues, we propose a hardware-algorithm co-design framework that combines a custom Spatially Varying Exposure (SVE) sensor with a physics-aware dehazing algorithm. The SVE sensor acquires multi-exposure data in a single shot, enabling robust haze assessment without relying on idealized atmospheric models. Our approach dynamically estimates haze density, performs region-adaptive illumination optimization, and applies multi-scale entropy-constrained fusion to effectively separate haze from scene radiance. Validated on real launch imagery and controlled experiments, the framework demonstrates superior performance in recovering physically accurate visual information of the plume and engine region. This offers a reliable image basis for extracting key mechanical parameters, including particle velocity, flow instability frequency, and structural vibration, thereby supporting precise quantitative analysis in extreme aerospace environments.
\end{abstract}


\section{Introduction}
\label{sec:introduction}
Optoelectronic imaging systems are indispensable in aerospace launch monitoring, enabling trajectory measurement, anomaly detection, and quantitative mechanical analysis through techniques such as particle image velocimetry, digital image correlation, and high-speed videography \cite{rocket,2014High,reconstruction,velocimetry,Gao23}. However, these measurements face severe challenges due to intense combustion-generated smoke and transient luminosity, which together produce extreme dynamic range and localized over-exposure. These conditions degrade image quality by obscuring texture patterns essential for Particle Image Velocimetry (PIV) and Digital Image Correlation (DIC), while over-exposure obliterates key features, compromising the accuracy of extracted parameters such as flow velocity, strain, and vibration. These issues highlight the critical need for imaging solutions capable of simultaneously handling dense smoke and high dynamic range illumination.

\begin{figure*}[htbp]
	\centering
	\includegraphics[width=0.95\textwidth]{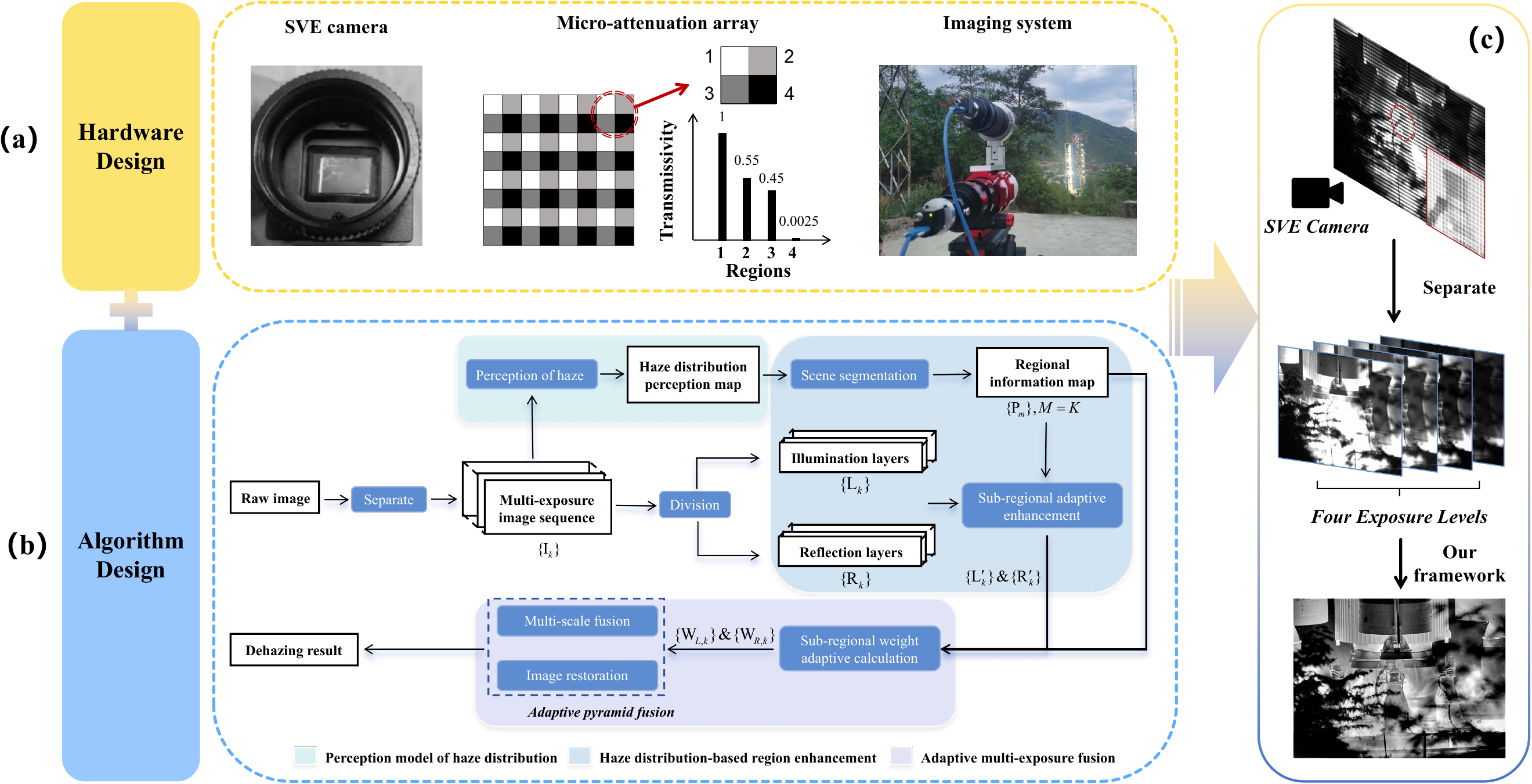}
	\caption{ Hardware-algorithm co-design framework diagram. (a) HDR imaging hardware design; (b) algorithm architecture with color-coded regions corresponding to chapters below; (c) overall processing flow of the framework.}
	\label{fig:MAIN}
\end{figure*}

To manage High Dynamic Range (HDR) imaging, Multiple Exposure Fusion (MEF) techniques are widely used. Merten \emph{et al.} \cite{2009Exposure} pioneered a pixel-based fusion method using weighted averages based on contrast, saturation, and well-exposedness, valued for its computational efficiency though often limited in preserving fine detail. Ma \emph{et al.} \cite{SPD-MEF} proposed Structural Patch Decomposition (SPD-MEF) to better preserve structural information, later improved by methods such as PESPD-MEF \cite{PESPD-MEF} through enhanced perceptual factors. More recently, learning-based methods \cite{2017DeepFuse, Detail-Refinement, U2D2Net, Network-Enabled, AIM-MEF} have substantially advanced the performance of MEF. Nevertheless, the domain-specific nature of launch scenarios and scarce training data limit deep learning's applicability here.

Image dehazing, which aims to reverse atmospheric scattering effects, remains a challenging and ill-posed problem \cite{ZHAO1998221, Removal2021, Non-Homogeneous}. The classic Dark Channel Prior (DCP) method \cite{DCP} estimates transmission maps to remove haze but often introduces halo artifacts near bright regions. Alternative cues such as saturation characteristics have also been explored—for instance, Ling \emph{et al.} \cite{SLP} introduced a saturation line prior that correlates saturation and brightness for haze mitigation. Despite these advances, methods based on hand-crafted priors are inherently constrained by underlying assumptions and lack generalizability \cite{artificial2018, Variational, Multi-exposure2024}.

Recognizing the interdependence of HDR and dehazing, several fusion-based approaches have attempted to integrate these tasks \cite{2018Multi-Exposure,NURIT2021103500}. Galdran \cite{artificial2018} proposed an artificial multi-exposure fusion technique using gamma correction on underexposed hazy images. Zhou \emph{et al.} \cite{PSD} developed a model-based method employing MEF to estimate haze distribution. A key limitation is treating dehazing as a separate pre- or post-processing step, failing to exploit the synergy between exposure and haze. Their dependence on fragile physical priors further limits use in extreme launch environments.

Although hardware-algorithm co-design is not new---exemplified by Nayar's seminal Spatially Varying Exposure (SVE) hardware \cite{SVE} for general HDR imaging and Galdran's software-based AMEF method \cite{artificial2018}---their applicability in supporting precise mechanics measurement in extreme launch scenarios remains limited. Nayar's SVE was not optimized for the radiometric profile of combustion haze, and AMEF relies on synthesizing exposures from a single image, which cannot accurately capture the complex scattering effects critical for physical accuracy \cite{aero-optical}. Such limitations fundamentally restrict their utility in providing the high-fidelity image data required for quantitative mechanical analysis under extreme conditions.

To address this critical gap, we propose a tightly coupled hardware–algorithm framework specifically designed for harsh launch environments. Our approach is built upon a customized SVE imaging sensor that captures spatially registered multi-exposure data in a single shot, directly recording the radiometric response of haze. Moving beyond model-based assumptions, we introduce a data-driven dehazing algorithm that dynamically estimates haze density from the SVE image statistics, avoiding reliance on idealized atmospheric models. The haze perception guides an adaptive fusion process combining region-aware photometric enhancement and entropy-constrained multi-scale reconstruction, ultimately recovering a clear, high dynamic range image that faithfully represents scene radiance. This provides a reliable image foundation for subsequent optical measurements to accurately extract mechanical parameters.
The main contributions of this work are summarized as follows:
\begin{itemize}
	\setlength{\itemsep}{3pt}
	\setlength{\parsep}{3pt}
	\setlength{\parskip}{3pt}
	\item[$\bullet$] \textbf{A specialized co-design framework:} We introduce an integrated system combining a custom SVE imaging system with a dedicated dehazing-fusion algorithm, specifically designed for extreme aerospace environments.
	\item[$\bullet$] \textbf{A statistical-driven haze perception model:} Unlike physical prior-based methods, our model dynamically infers haze distribution from multi-exposure statistics, enhancing adaptability in complex scenes.
	\item[$\bullet$] \textbf{An adaptive sub-region fusion strategy:} We develop a haze-aware segmentation and weighting mechanism that enables localized enhancement and artifact-free fusion, outperforming global fusion approaches.
\end{itemize}

\section{Proposed Framework}
The proposed framework, outlined in Figure \ref{fig:MAIN}, combines specialized cameras with a novel processing algorithm. The algorithm is structured around three core modules: haze distribution perception (green), region-based perceptual enhancement (blue), and adaptive multi-exposure fusion (purple). Owing to the intricate nature of rocket combustion, we make no distinction between fog and smoke, unified under the term "haze." The overarching goal is to achieve significant visibility enhancement through advanced image processing, as elaborated in the following subsections.

\subsection{Spatially Varying Exposure Camera}
The SVE imaging paradigm, introduced by Nayar \emph{et al.} \cite{SVE}, offers a transformative approach for HDR imaging by spatially modulating photosensitivity. This innovation addresses the limitations of traditional multi-exposure fusion techniques, enabling the simultaneous capture of scene radiance variations across different sensor subregions in a single exposure window. This approach effectively eliminates motion artifacts often introduced by multi-frame synthesis in dynamic scenes. Inspired by the adaptive, localized photosensitive mechanisms in natural visual systems, such as the human retina, SVE has been further optimized for exposure precision and energy conversion efficiency through advanced sensor architectures \cite{IROS2022,Spatially}.

To meet the demanding requirements of rocket plume visualization under extreme launch conditions, we develop a customized SVE imaging system based on the foundational architecture in Ref.~\citenum{SVE}. Unlike general-purpose SVE designs, our system is specifically co-engineered to address the opto-radiometric challenges of launch vehicle monitoring, especially the intense scattering and emission characteristics of combustion haze. Through targeted modifications to industrial-grade image sensor architectures and readout circuitry, the system is optimized to capture the distinctive radiometric and scattering signatures of combustion haze across multiple exposures. This transition from a general theoretical model to a domain-specific imaging apparatus forms the hardware cornerstone of our framework. It ensures the acquisition of high-fidelity data essential for subsequent physics-aware processing.

Compared to other real-time HDR imaging systems \cite{varying,2016Real,2016HDR,Spikes}, our SVE camera offers distinct practical advantages: it is compact, portable, easy to deploy, and inherently free of parallax artifacts between differently exposed sub-images, eliminating the need for additional alignment. These characteristics make it particularly suitable for distributed multi-sensor configurations in large-scale launch site monitoring.

The hardware core is an integrated attenuator array that acts as a spatially varying optical filter. Each 2$ \times $2 macro-pixel comprises four micro-attenuators with defined transmittances (1:0.55:0.45:0.0025), mapped to individual CMOS pixels (Figure \ref{fig:MAIN}(a)). This ratio set is a co-design outcome, driven by the extreme radiometry of rocket launches and the data needs of our dehazing algorithm. It brackets a $>$120 dB range to capture dark details and avoid plume saturation, provides closely-spaced exposures to sample complex plume-smoke gradients for haze perception, and ensures that scene radiance is optimally sampled within the high-sensitivity region of the sensor's non-linear response function.

The camera's output is a single raw image $I_{raw}$, intrinsically containing a mosaic of the four different exposures. The first processing step demultiplexes this mosaic into a sequence $\{I_k\}_{k=1}^K$ (with $K=4$) of low-resolution sub-images, each representing one exposure level. The extraction of a pixel $I_k(u, v)$ from the raw image is governed by:
\begin{equation}
	I_k(u, v) = I_{raw}(2u - 1 + \delta_r, 2v - 1 + \delta_c),
\end{equation}
where $(u, v)$ are coordinates in the sub-image. The row offset $\delta_r = \lfloor (k-1)/2 \rfloor$ and column offset $\delta_c = (k-1) \bmod 2$ map these coordinates back to the raw image grid. At this stage, each $I_k$ has one quarter of the native sensor resolution (e.g., 1224$ \times $1024 from 2448$ \times $2048).

While the SVE architecture thus involves a theoretical reduction in spatial sampling due to its 2$ \times $2 macro-pixel design, this is effectively mitigated by a dedicated reconstruction stage. Analogous to demosaicing, we apply an edge-preserving, gradient-corrected interpolation algorithm \cite{Malvar2004,IROS2022} to each exposure level, yielding a set of full-resolution images $\{\tilde I_k\}_{k=1}^K$ that are spatially registered and exposure-varying. Critically, for quantitative measurements such as PIV/DIC, this constitutes a strategic trade-off: by preventing saturation in the plume core at the hardware level, the system preserves usable high-contrast texture across the entire field of view—information that would be irretrievably lost in conventional imaging—while the reconstruction step restores sub-pixel spatial accuracy.

In the remainder of this paper, all algorithmic modules (haze perception, region-based enhancement, and multi-exposure fusion) operate on the reconstructed full-resolution sequence $\{\tilde I_k\}$, which we denote generically as $\{I_k\}$ for notational simplicity.

\begin{figure}[tp]
	\centering		
	{\includegraphics[width=0.6\textwidth]{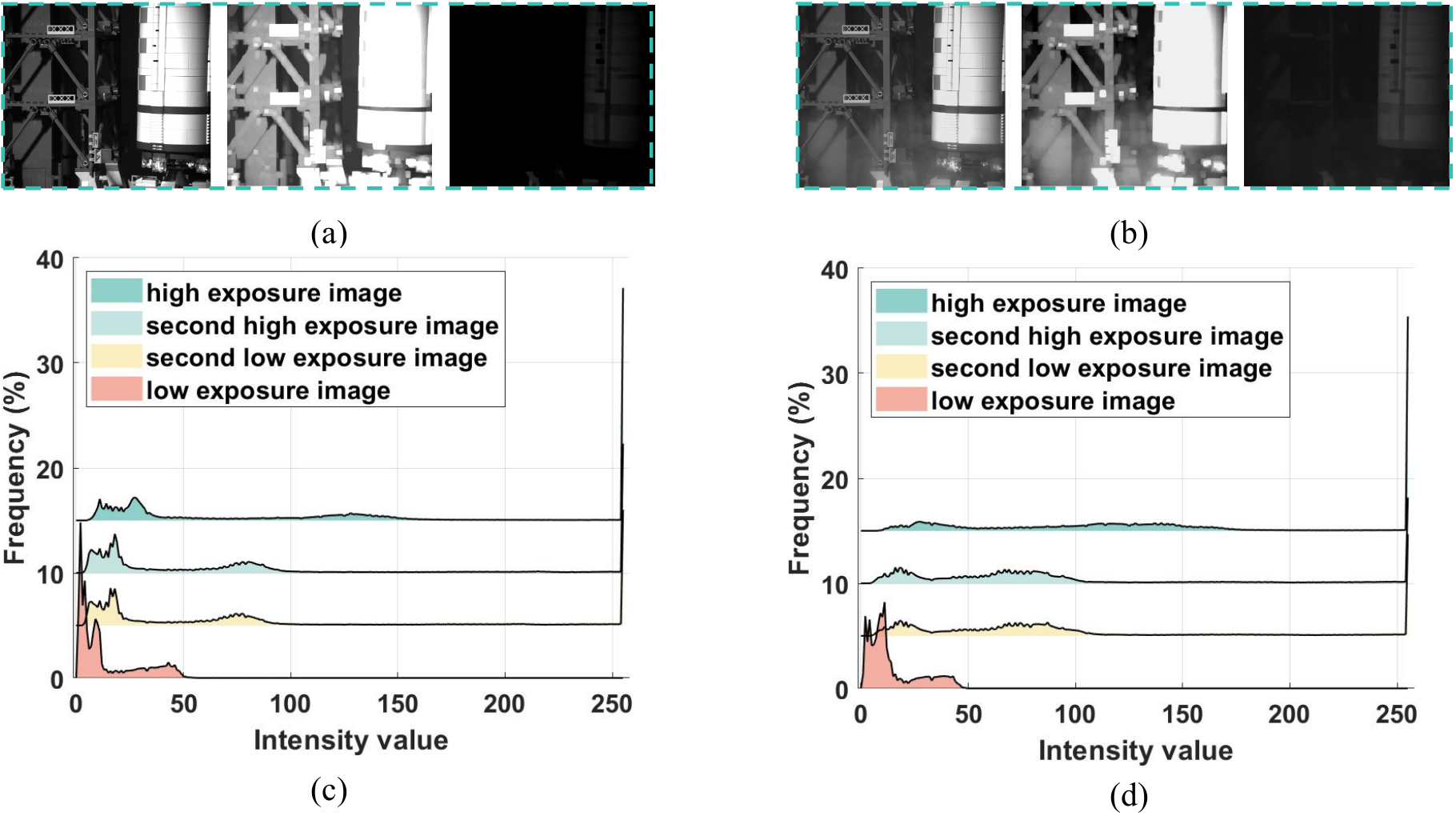}}
	\caption{Channel characteristics in multi-exposure imaging: (a) Representative haze-free scene image with corresponding bright and dark channels; (b) Representative hazy scene image with corresponding bright and dark channels; (c) Grayscale histograms for the four distinct exposure levels of the SVE camera for the scene in (a); (d) Grayscale histograms for the four distinct exposure levels of the SVE camera for the scene in (b).}
	\label{fig:DCP-BCP}
	\vspace{0.2cm}
	\centering		
	{\includegraphics[width=0.6\textwidth]{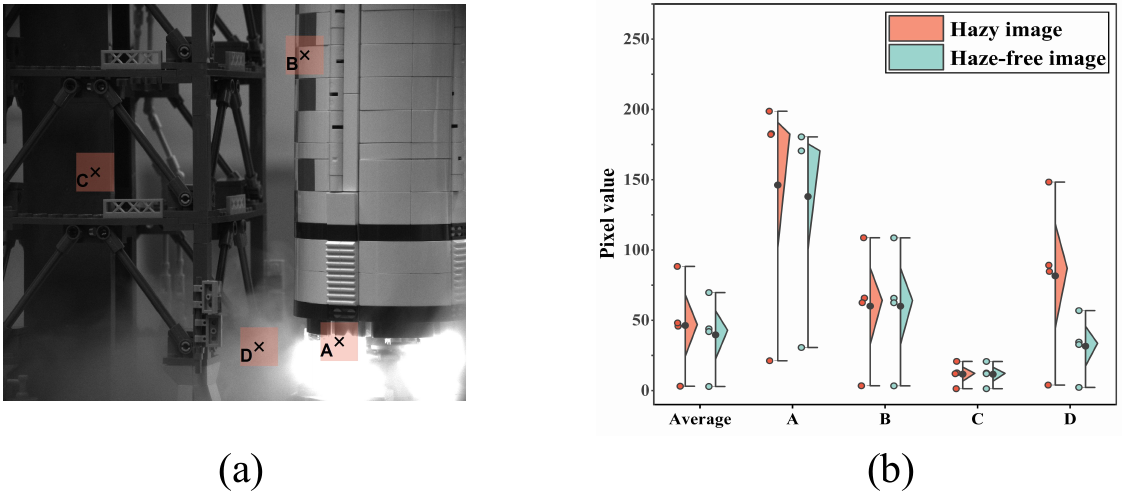}}
	\caption{Statistical features of brightness variation in image patches: (a) selected patch; (b) boxplot of brightness variation across different exposures. The hollow circle indicates the mean value at different exposure levels, while the solid circle represents the overall mean.}
	\label{fig:var}
\end{figure}

\subsection{Haze Distribution Perception}
This section presents our data-driven haze perception model, which utilizes the distinctive multi-exposure capability of the SVE sensor to characterize haze without depending on idealized physical priors. Under extreme launch conditions, classical assumptions such as the Dark Channel Prior (DCP) are frequently ineffective, as bright haze-like interferences often violate their underlying principles. Our approach overcomes these constraints by directly learning the statistical properties of haze from radiometric data across multiple exposures, thereby establishing a more robust relationship between observed intensity and scattering density.

\subsubsection{Feature Extraction from Single-Exposure Images}
In rocket launch observation scenarios, atmospheric haze exhibits intensity amplification under complex illumination, frequently leading to localized overexposure—a key challenge for photometric measurement. Consequently, hazy regions demonstrate significantly elevated luminance values compared to the global image mean. 

To estimate the haze factor, the mean brightness information ($BI$) is first calculated across the exposure sequence:
\begin{equation}
	BI(x) = \frac{1}{K}\sqrt {\sum\limits_{k = 1}^K {(max(} {{\rm{I}}_k}(x),T) - {\mu _k}{)^2}}.
	\label{eq:BI}
\end{equation}

Let $k$ index exposures, ${{\rm{I}}_k}\left( x \right)$ denotes the grayscale intensity at pixel $x$ in the $k$-th exposure, and ${{\mu _k}}$ represents the mean intensity of frame $k$. A luminance threshold $T = {\mu _k}/2$ is implemented to eliminate low-intensity artifacts.

To enhance haze region identification, we use Weber contrast ($WC$) analysis to quantify contrast degradation, a direct consequence of scattering-induced light diffusion.
\begin{equation}
	WC(x) = \frac{1}{K}\sum\limits_{k = 1}^K {\frac{{|\nabla {{\rm{I}}_k}(x)|}}{{{{\rm{I}}_k}(x) + 1}}},
	\label{eq:WC}
\end{equation}
where $\nabla $ represents the gradient operator. In regions influenced by haze, the image brightness often shows consistently high gray levels with reduced gradients, leading to a characteristically reduced $WC$. 

\subsubsection{Radiometric Feature Analysis Across Exposures}
In this section, we enhance the characterization of haze by incorporating advanced photometric priors that are more resilient to the complex light field. Specifically, we combine the Dark Channel Prior (DCP) \cite{DCP} and the Bright Channel Prior (BCP) \cite{Wang2013,2017ICIVC} to extract Contrast Features (CF). Building on the proven effectiveness of DCP in grayscale domains for scattering media analysis \cite{PSD}, we extend this concept to multi-exposure data. As illustrated in Figure \ref{fig:DCP-BCP}, haze typically reduces the intensity of the bright channel, increases the intensity of the dark channel, smooths histograms, and suppresses extreme pixel values. These statistical shifts provide a physical basis for haze detection.

The atmospheric scattering model \cite{Optic,flows,Non-local} provides the fundamental physical model describing haze formation:
\begin{equation}
	{\rm{I}}(x) = {\rm{J}}(x)t(x) + {\rm{A}}(1 - t(x)),
	\label{eq:Model}
\end{equation}
where ${\rm{J}}(x)$ is the latent haze-free image, $t(x)$ is the medium transmission map, and $\rm{A}$ is the atmospheric light intensity. Within a local block, the transmission $t$ and atmospheric light ${\rm{A}}$ are assumed to be constant. This assumption allows for the simultaneous calculation of the bright and dark channels:
\begin{equation}
	\left\{ \begin{array}{l}
		{{\rm{I}}^{\rm{d}}}(x) = t \cdot {{\rm{J}}^{\rm{d}}}(x) + {\rm{A}}(1 - t)\\
		{{\rm{I}}^{\rm{b}}}(x) = t \cdot {{\rm{J}}^{\rm{b}}}(x) + {\rm{A}}(1 - t),
	\end{array} \right.
	\label{eq:Jd_Jb}
\end{equation}
where the dark and bright channels are derived by applying the minimum and maximum operators across four exposure levels.
Following \cite{atmos}, the contrast feature operator is derived through channel interactions. As ${{\rm{J}}^d}(x) \to 0$, the relationship can be expressed as:
\begin{equation}
	{{\rm{I}}^b}(x) = {{\rm{J}}^b}(x) \cdot (1 - \frac{{{{\rm{I}}^d}(x)}}{{\rm{A}}}) + {{\rm{I}}^d}(x),
\end{equation}
yielding the contrast relationship:
\begin{equation}
	{\frac{{{{\rm{J}}^b}(x)}}{{{{\rm{I}}^b}(x)}} = \frac{{1 - \frac{{{{\rm{I}}^d}(x)}}{{{{\rm{I}}^b}(x)}}}}{{1 - \frac{{{{\rm{I}}^d}(x)}}{{\rm{A}}}}} = \frac{{CF(x)}}{t}}.
\end{equation}
Thus, the contrast feature $CF(x)$ is defined as:
\begin{equation}
	{CF(x) = 1 - \frac{{{{\rm{I}}^d}(x)}}{{\max ({{\rm{I}}^b}(x),1)}}}.
	\label{eq:CF}
\end{equation}

By incorporating multi-exposure channels, the $CF$ feature extends the principles of the BCP and DCP priors, enabling more effective haze detection in dynamic sequences by capturing scattering variations across intensity levels.

Further discriminative capability is achieved by analyzing exposure-dependent variance fluctuations. As shown in Figure \ref{fig:var}, regions affected by haze (e.g., regions A and D) display variance magnitudes that significantly exceed the average variance of the entire image. This increase in variance serves as a key indicator for distinguishing haze-affected areas from non-haze regions.

Notably, even in haze-free images, region A may display significant variance changes due to local phenomena such as firelight fluctuations. However, haze further exacerbates these fluctuations, leading to more pronounced variance deviations. By leveraging this characteristic, haze-affected regions can be effectively identified through the analysis of pixel variance fluctuations across different exposure levels.

To quantify the radiometric instability caused by scattering, we analyze fluctuations in variance across different exposure levels. The normalized variance metric ${{\rm{Var}}(x)}$ is defined as follows:
\begin{equation} 
	{V(x) = \frac{{{\rm{Var}}(x) - \bar \chi }}{{\bar \chi  + \varepsilon }}},
	\label{eq:V}
\end{equation}
where ${{\rm{Var}}(x)}$ represents the cross-exposure variance, ${\bar \chi }$ is the geometric mean variance of the image, and $\varepsilon {\rm{ = 1}}{{\rm{0}}^{{\rm{ - 6}}}}$ is introduced to prevent singularity.

\begin{figure}[tp]
	\centering		
	{\includegraphics[width=0.6\textwidth]{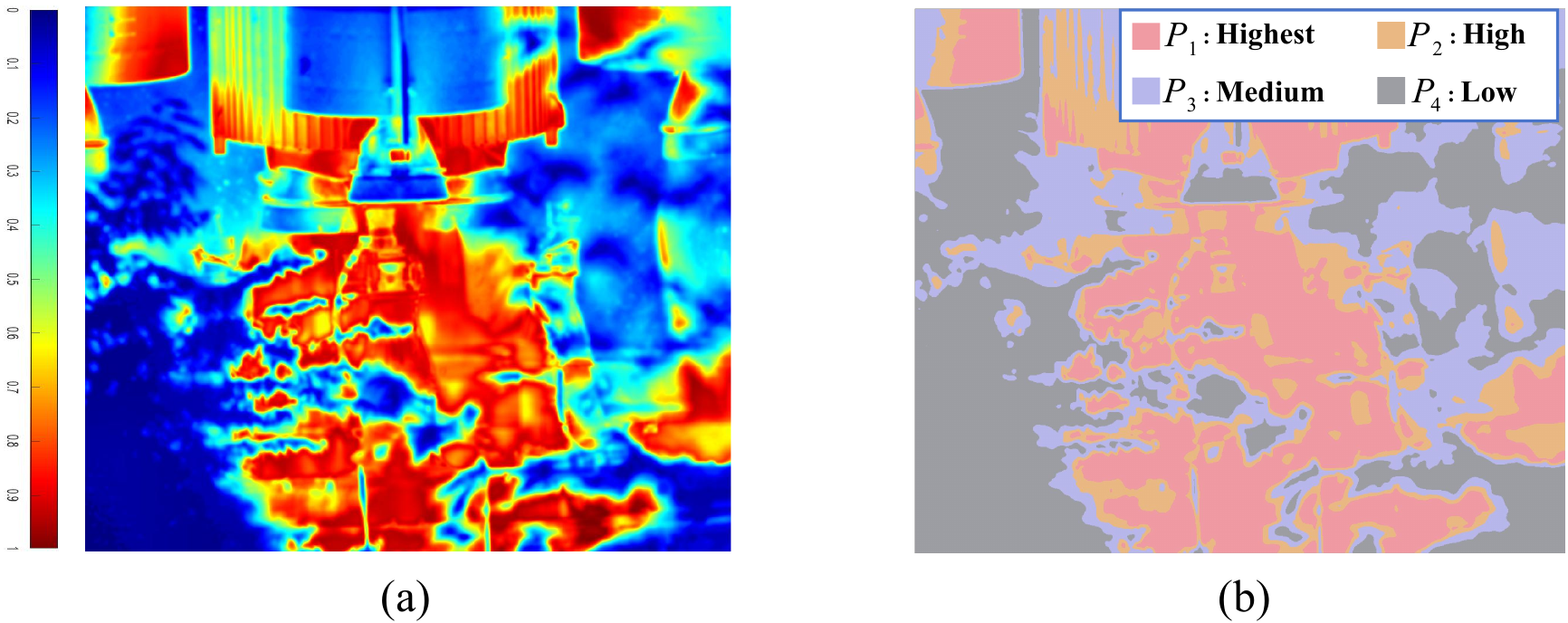}}
	\caption{Haze perception and segmentation results: (a) Probabilistic haze distribution map with color bar indicating normalized haze density, increasing with haze concentration; (b) scene-adaptive region segmentation ($m$ = 4), with regions labeled by decreasing haze probability (${p_1}$ to ${p_4}$).}
	\label{fig:haze}
\end{figure}

Feature normalization employs exponential mapping to compress the dynamic range for stable fusion:
\begin{equation}
	f'(x) = \exp (\frac{{ - f{{(x)}^2}}}{{\rm{c}}}),
	\label{eq:mapping function}
\end{equation}
with ${\rm{c}}$ controlling dynamic range compression. Each of the calculated features ($BI(x)$, $WC(x)$, $CF(x)$, and $V(x)$) is normalized using the mapping described in Eq.(\ref{eq:mapping function}) before combination.
The integrated haze perception model combines four key discriminative features:
\begin{equation}
	{\rm{F}}(x) = \alpha BI(x) + \beta WC(x) + \gamma CF(x) + \sigma V(x),
	\label{eq:model}
\end{equation}
where the coefficients, optimized empirically on our launch datasets, are set as $(\alpha ,\beta ,\gamma ,\sigma ) = (0.2,0.2,0.3,0.3)$ to balance detection sensitivity and specificity, while minimizing false alarms in bright regions. Figure \ref{fig:haze}(a) illustrates the resulting haze distribution model, emphasizing areas with a higher likelihood of haze occurrence. This model effectively converts captured radiometric data into a spatially varying measure of scattering density, which serves as the physical basis for implementing sub-region-specific enhancements.

\subsection{Perceptual Region-Based Enhancement} 
This section introduces a sub-region approach that implements customized processing strategies tailored to each region. The methodology combines localized gamma correction with global contrast preservation mechanisms to address non-uniform atmospheric scattering effects. This dual-scale optimization strategy effectively reduces haze density variations while preserving photometric consistency.

\subsubsection{Scene Segmentation} 
Moving beyond conventional semantic segmentation \cite{2019Scene}, our approach utilizes the physically-derived haze distribution perception map ${\rm{F}}(x)$ for scene partitioning. This method focuses on identifying regions exhibiting distinct levels of scattering density, thereby enabling physically-grounded sub-regional analysis and processing

The haze distribution perception map ${\rm{F}}$ is partitioned into $M$ equal luminance intervals, with each region ${{\rm{P}}_m}$ defined as the set of pixels satisfying the following condition:
\begin{equation}
	{{\rm{P}}_m} = \{ x|{\theta _m} \le {\rm{F}}(x) \le {\theta _{m + 1}}\},
\end{equation}
\begin{small}
	\begin{equation}
		{\theta _m} = \frac{{M - m + 1}}{M}\left( {\max {\rm{F}}(x) - \min {\rm{F}}(x)} \right) + \min {\rm{F}}(x).
	\end{equation}
\end{small}
Let ${{\rm{P}}_m}$ represent the pixel set of the $m$-th region, and $M$ the total number of regions. By adopting parametric coupling with $M = K = 4$, we ensure index consistency between the segmented regions (indexed by $m$) and the input exposure sequence (indexed by $k$), as introduced in \cite{2019Scene}. This choice of four regions directly corresponds to the four SVE exposure levels, balancing detail capture for haze distribution with computational efficiency for real-time processing.

As shown in Figure \ref{fig:haze}(b), the haze distribution perception model effectively detects regions with varying haze concentrations, with higher values indicating a greater likelihood of haze presence. This segmentation approach provides a robust foundation for subsequent sub-regional enhancement, enabling targeted processing strategies that maximize image quality improvements in haze-affected areas.

\subsubsection{Sub-Regional Adaptive Enhancement} 
This subsection presents a spatially adaptive enhancement framework based on the photometric analysis of partitioned image regions. The process begins by extracting photometric components through weighted guided image filtering (WGIF) \cite{WGIF}, a method known for its effectiveness in preserving edge sharpness while suppressing noise. Using this technique, the input image ${\rm{I}}(x)$ is decomposed into an illumination layer ${\rm{L}}(x)$ and a reflectance layer ${\rm{R}}(x)$ according to the Retinex-based model \cite{Water-related, Simultaneous}. This model separates the overall light field distribution from the inherent reflectivity of scene objects, thereby establishing a tractable foundation for targeted photometric corrections. The decomposition is formally expressed as:
\begin{equation}
	{\rm{I}}(x) = {\rm{L}}(x) \cdot {\rm{R}}(x).
\end{equation}

\begin{figure*}[tp]
	\centering
	\includegraphics[width=0.95\textwidth]{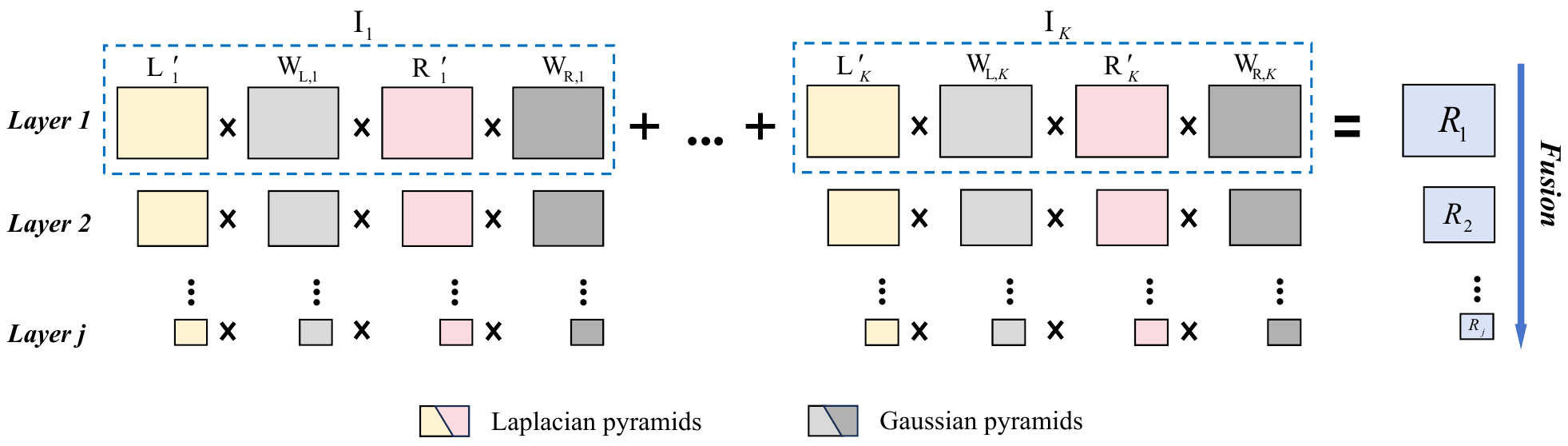}
	\caption{The general pipeline of adaptive pyramid fusion: The illumination and reflection layers are decomposed into Laplacian pyramids, while the weighting maps are decomposed into Gaussian pyramids. The symbol $\times $ denotes the dot product. ${R_1}{\rm{  -  }}{R_j}$ represent the layers of the resulting Laplacian pyramid.}
	\label{fig:pyramid}
\end{figure*}

To adaptively adjust the photometric response, we compute the geometric mean luminance within each sub-region ${{\rm{P}}_m}$. This measure offers a robust estimate of regional brightness, particularly under hazy conditions. The geometric mean is defined as: 
\begin{equation}
	G({\rm{L}}|{{\rm{P}}_m}) = \exp \left( {\frac{1}{{|{{\rm{P}}_m}|}}\sum\limits_{x \in {{\rm{P}}_m}} {\log } \left( {\max \left( {{\rm{L}}(x),\varepsilon } \right)} \right)} \right),
\end{equation}
where a small constant $\varepsilon$ is introduced to avoid numerical instability. This value then guides an adaptive gamma correction scheme, formulated as:
\begin{equation}
	{\gamma _{{k}}} = {\left( {2 + G({{\rm{L}}_{{k}}}|{{\rm{P}}_m})} \right)^{(2 \cdot G({{\rm{L}}_{{k}}}|{{\rm{P}}_m}) - {\zeta _{{k}}})}},
\end{equation}
with ${{k}} = K - m + 1$ indexing exposure levels ($K$: total exposure count). The parameter ${\zeta _{{k}}}$ corresponds to the standard brightness for each region, while the corrected luminance for each pixel is expressed as ${{\rm{L'}}_k}(x) = {{\rm{L}}_k}{(x)^{{\gamma _k}}}$.

Finally, the reflection layer, which captures high-frequency details, is enhanced using WGIF \cite{WGIF} to suppress noise and enhance edges, resulting in a processed reflection layer $\{{{\rm{R'}}_k}\}$. This enhancement strategy ensures that both the illumination and reflectance components are optimized for clarity and detail, thereby significantly improving the overall image quality in regions affected by haze.

\subsection{Adaptive Multi-Exposure Fusion} 
\subsubsection{Sub-Regional Weight Estimation}
Fusion weights are computed independently for the illumination and reflectance components. This dual-channel weighting strategy is designed to optimally preserve the global brightness distribution (via $L$) and high-frequency detail information (via $R$) during fusion, while effectively suppressing noise and artifacts introduced by scattering.

For the illumination layer, we develop adaptive weights based on photometric analysis approaches introduced in \cite{ICIP,2009Exposure}, incorporating both regional statistics and cumulative histogram properties. The weight calculation employs gradient analysis of the cumulative histogram, emphasizing pixels with smoother distribution transitions as these are considered optimally exposed and more suitable for fusion. The weight function for the illumination layer is defined as:
\begin{equation}
	{{\rm{W}}_{{{\rm{L}}_1},k}}(x) = \frac{{Gra{d_k}{{({{{\rm{L'}}}_k}(x))}^{ - 1}}}}{{\sum\limits_{k = 1}^K G ra{d_k}{{({{{\rm{L'}}}_k}(x))}^{ - 1}} + \varepsilon }},
\end{equation}
where $Gra{d_k}({{\rm{L'}}_k}(x))$ represents the gradient of the cumulative histogram at intensity ${{\rm{L'}}_k}(x)$. 

The weight estimation is further refined through Gaussian mixture modeling:
\begin{equation}
	{{\rm{W}}_{{{\rm{L}}_2},k}}(x) = \frac{1}{\psi }\exp \left( { - \frac{{{{({{{\rm{L'}}}_k}(x) - {u_k})}^2}}}{{2{\sigma ^2}}}} \right),
\end{equation}
where ${\psi }$ denotes the normalized coefficient and parameter ${{\sigma }}$ controls distribution spread, and the regional mean ${{u_k}}$ is adjusted based on the average brightness of each region.
\begin{equation}
	\begin{array}{*{20}{c}}
		{{u_k} = G({{\rm{L}}_k}|{{\rm{P}}_m}),}&{k = K - m + 1}.
	\end{array}
\end{equation} 

The composite illumination weight $\{ {{\rm{W}}_{{\rm{L}},k}}\} $ is obtained through multiplicative fusion:
\begin{equation}
	{{\rm{W}}_{{\rm{L}},k}} = {{\rm{W}}_{{{\rm{L}}_1},k}} \cdot {{\rm{W}}_{{{\rm{L}}_2},k}}.
\end{equation} 
This dual-weight mechanism effectively attenuates haze interference during the fusion process, ensuring that the illumination layer is optimally enhanced.

For the reflectance component, edge-aware weights are generated by evaluating the relative edge strength of a central pixel against its local neighborhood:
\begin{equation}
	{{\rm{W}}_{{\rm{R}},k}} = \frac{1}{N}\sum\limits_{x = 1}^N {\frac{{\sigma _{{{\rm{R}}_k},3}^2(x') + \varepsilon }}{{\sigma _{{{\rm{R}}_k},3}^2(x) + \varepsilon }}} ,
\end{equation}
where $\varepsilon$ is a small constant that depends on the dynamic range of the input, $N$ denotes the total number of pixels, and $\sigma _{{{\rm{R}}_k},3}^2(x')$ represents the variance of ${{\rm{R}}_k}$ over a $3 \times 3$ window centered at the corresponding location.

\subsubsection{Adaptive Pyramid Fusion}
To reduce fusion artifacts caused by discontinuous weight distributions \cite{Detail-Enhanced,Multiscale}---commonly encountered near haze boundaries and other non-smooth regions---we introduce an adaptive pyramid fusion framework based on multi-scale decomposition. As shown in Figure \ref{fig:pyramid}, the method employs a joint Gaussian–Laplacian pyramid to seamlessly integrate illumination and reflectance components across scales.

The decomposition depth $j$ is determined by:
\begin{equation}
	j = \left\lfloor {lo{g_2}(min(h,w))} \right\rfloor  - 2,
\end{equation}
where $h \times w$ is the image size, and $\left\lfloor \cdot \right\rfloor$ denotes the floor function. 
The Gaussian pyramid is constructed recursively as:
\begin{equation}
	{G^{(l + 1)}} = Down({G^{(l)}} \otimes {G_{kernel}}),
	\label{eq:guass}
\end{equation}
with ${G_{kernel}}$ representing the Gaussian kernel, and ${Down}$ represents downsampling.

The Laplacian pyramid captures high-frequency details through the difference:
\begin{equation}
	{L^{(l)}} = {G^{(l)}} - Up({G^{^{(l + 1)}}}) \otimes {G_{kernel}},
	\label{eq:laplacian}
\end{equation}
where ${Up}$ denotes upsampling and $\otimes$ indicates convolution.
This multi-scale strategy enables seamless fusion: large-scale illumination variations are blended at coarse levels, while fine details are preserved at finer levels.

For each scale, the fusion operation is defined as:
\begin{footnotesize}
	\begin{equation}
		{R_{\rm{k}}}^{(l)} = \{ {L^{(l)}}\{ {{{\rm{L'}}}_k}\}  \cdot {G^{(l)}}{\rm{\{ }}{{\rm{W}}_{{\rm{L}},k}}\} \} \{ {L^{(l)}}\{ {{{\rm{R'}}}_k}\}  \cdot {G^{(l)}}{\rm{\{ }}{{\rm{W}}_{{\rm{R}},k}}\} \}, 
	\end{equation}
\end{footnotesize}
where $l = 1,2, \ldots ,j$ denotes the pyramid level, and $k$ indexes the exposure levels. This operation combines the high-frequency details and weighted contributions from each exposure level to produce the final fused image.

The final dehazing result is generated by reconstructing the Laplacian pyramid composed of ${R^{(l)}}$. 
Each layer of the final Laplacian pyramid is obtained as:
\begin{equation}
	\begin{array}{*{20}{c}}
		{{R^{(l)}} = \sum\limits_{k = 1}^K {{R_k}^{(l)},} }&{l = 1,2, \ldots ,j}.
	\end{array}
\end{equation}

As shown in Figure \ref{fig:MAIN}(c), the proposed framework effectively reduces haze while preserving photometric naturalness. The adaptive pyramid fusion strategy effectively balances global photometric consistency and local textural detail. This ensures the final output retains natural visual appearance while maximizing the visibility of critical features in the scene, which is paramount for subsequent monitoring and analysis tasks.

\begin{figure*}[t]
	\centering		
	\includegraphics[width=0.68\textwidth]{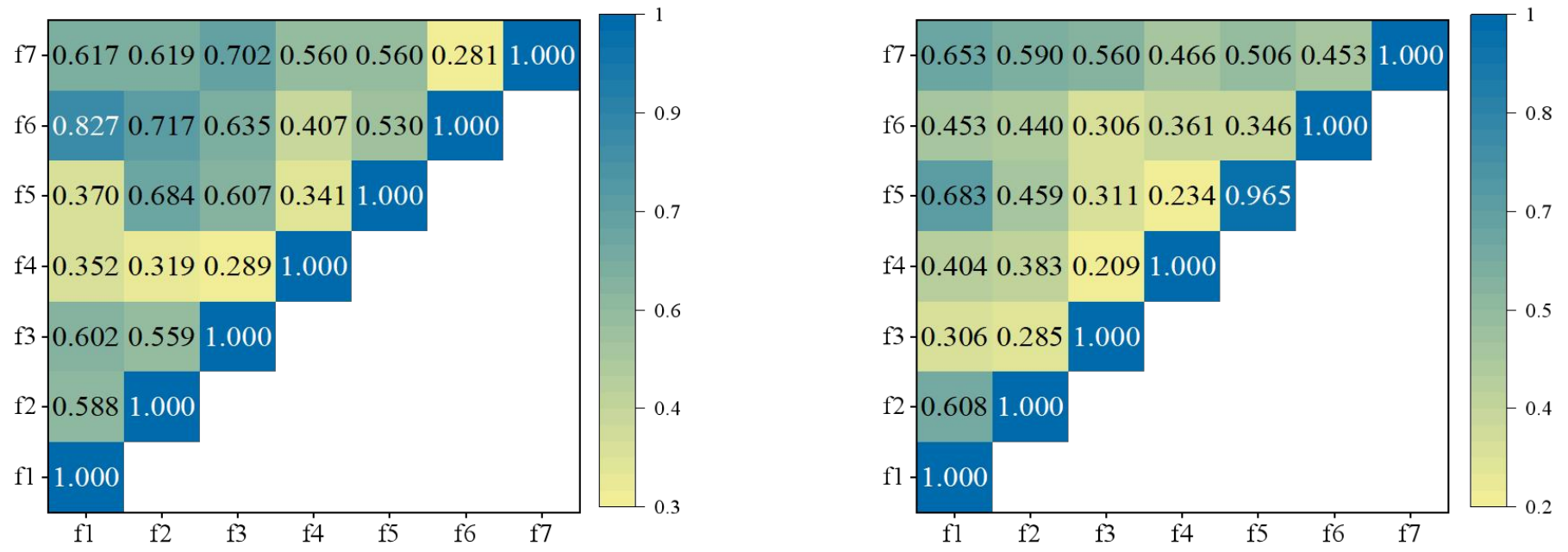}
	\caption{The correlation coefficient maps between features are calculated using (a) PCC and (b) MIC.}
	\label{fig:hotmap}
	\vspace{0.2cm}
	\centering		
	\includegraphics[width=0.76\textwidth]{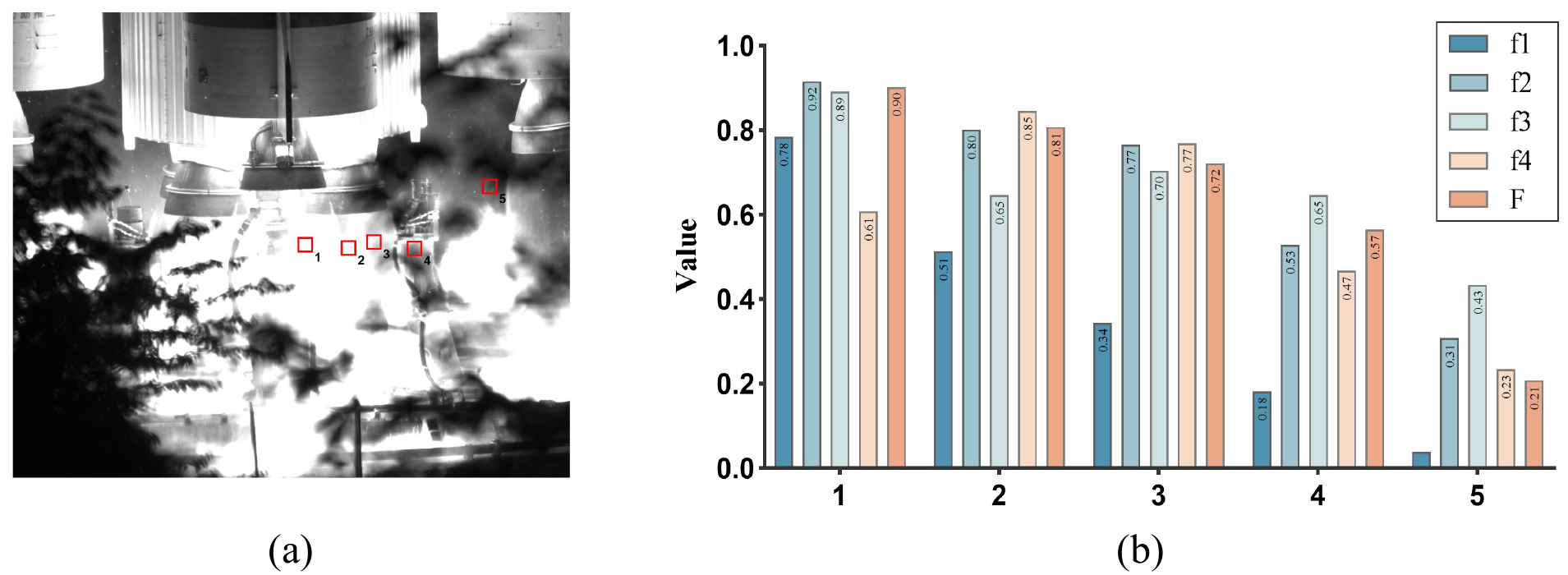}
	\caption{Image block value under different feature operators. (a) Haze image. (b) The metrics of five image patches shown in the hazy image (a).}
	\label{fig:score}
\end{figure*}	

\section{Experiments}
In this section, a series of experiments are conducted to comprehensively assess the performance of the proposed dehazing framework. Given that the framework inherently combines multi-exposure fusion with haze suppression, the subsequent experiments are designed to evaluate its efficacy from these two critical viewpoints.

\subsection{Implementation Details}
Due to the specialized nature of the SVE camera-based methodology, conventional benchmark datasets are not suitable for validating performance under the unique opto-radiometric conditions of rocket launches \cite{IROS2022}. Therefore, we construct several dedicated datasets to rigorously evaluate our framework:

\emph{1) Dataset 1:} On-Site Launch Sequences. This dataset comprises 475 image sequences captured on-site during an actual rocket launch using the SVE camera. The high resolution (2448 $\times$ 2048) captures detailed dynamics of plume evolution and combustion haze, providing a ground-truth benchmark for extreme-environment performance.

\emph{2) Dataset 2:} Laboratory Combustion. To enable controlled validation, this dataset consists of 560 images of fuel combustion processes captured in a laboratory setting. It allows for the analysis of haze and luminosity characteristics under repeatable conditions.

\emph{3) Dataset 3:} Simulated Haze with Ground Truth. This dataset includes 512 images generated in a laboratory simulation chamber, featuring a rocket model and a haze generator. It provides paired hazy and haze-free reference images, enabling full-reference quantitative analysis.

\begin{figure*}[t]
	\centering	
	{\includegraphics[width=0.92\textwidth]{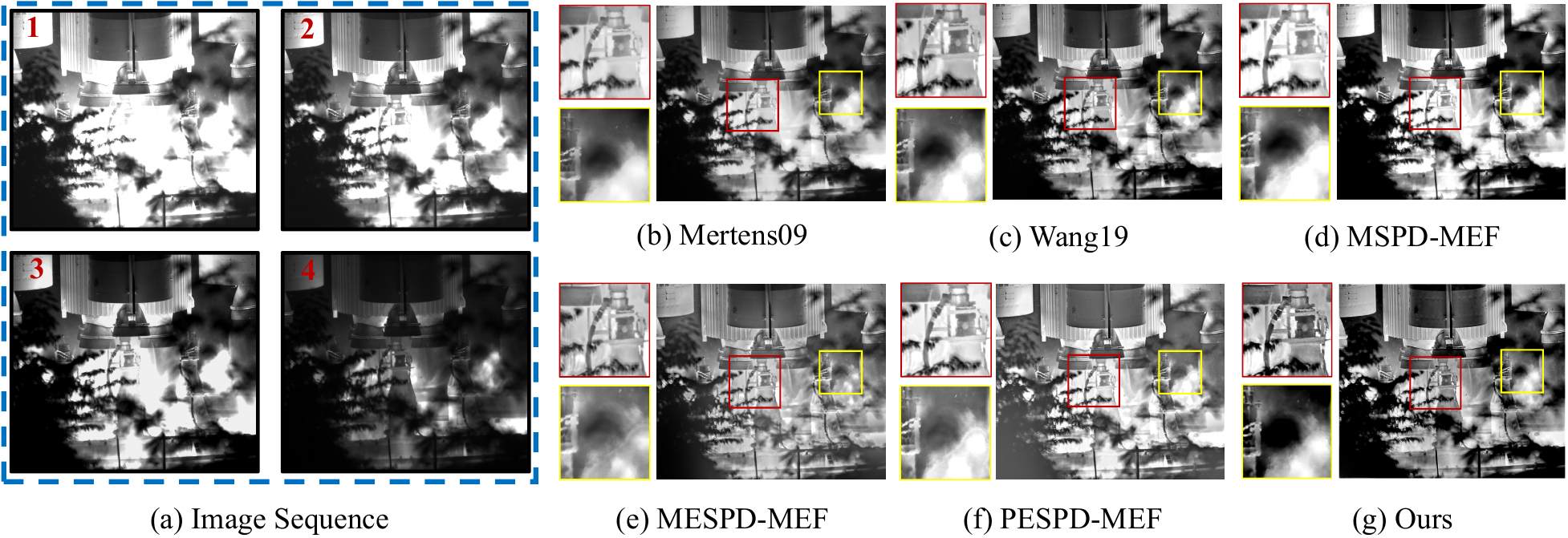}}
	\caption{Visual results of various multi-exposure methods on \emph{Dataset1}. (a) Input sequence. (b) to (g) show the results reconstructed using different methods: Mertens09 \cite{2009Exposure}, Wang19 \cite{Detail-Enhanced}, MSPD-MEF \cite{MSPD-MEF}, MESPD-MEF \cite{MESPD-MEF}, PESPD-MEF \cite{PESPD-MEF}, and ours.}
	\label{fig:multi_result01}	
	\vspace{0.2cm}
	\centering	
	{\includegraphics[width=0.70\textwidth]{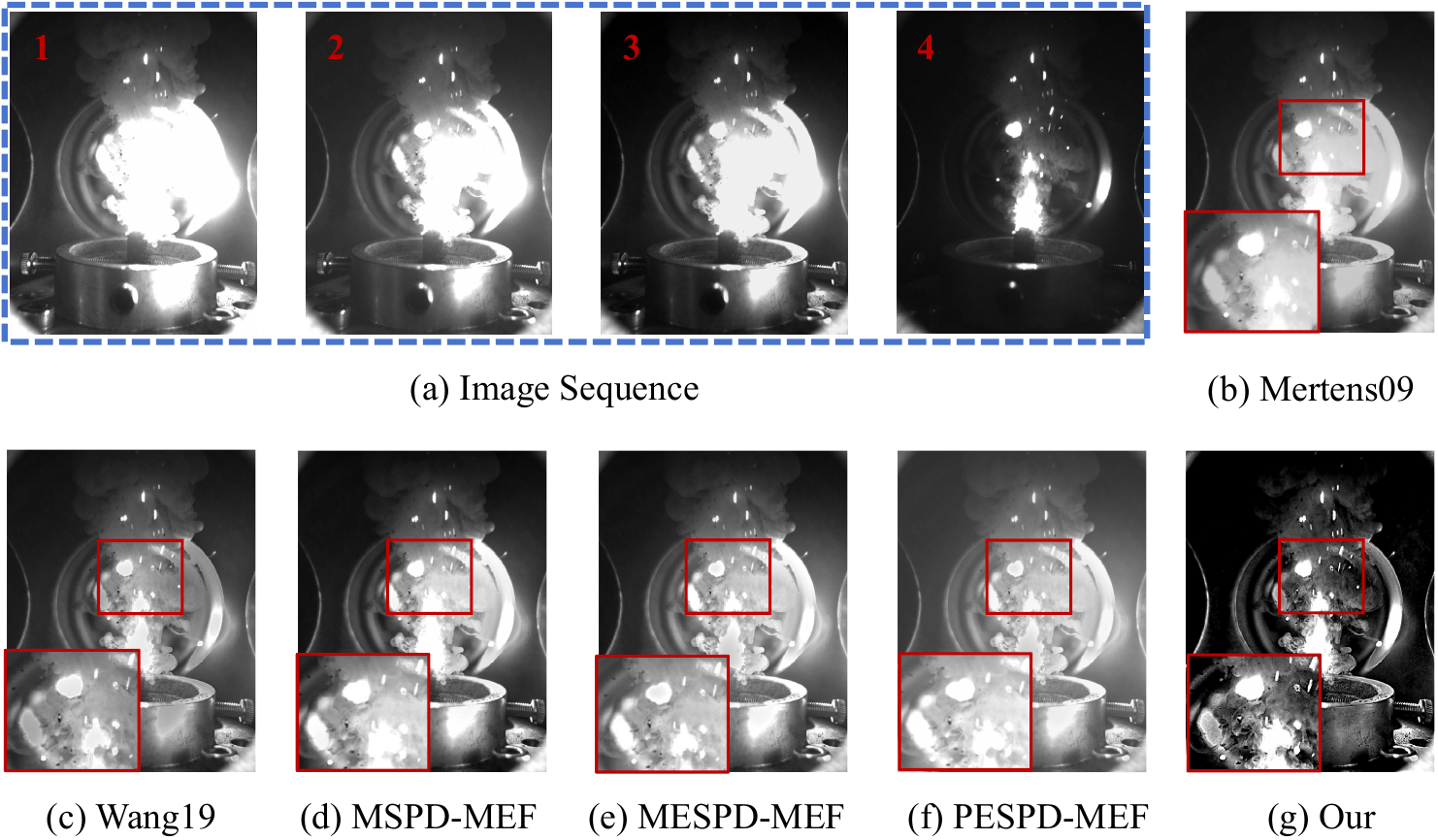}}
	\caption{Visual results of various multi-exposure methods on \emph{Dataset2}. (a) Input sequence. (b) to (g) show the results reconstructed using different methods: Mertens09 \cite{2009Exposure}, Wang19 \cite{Detail-Enhanced}, MSPD-MEF \cite{MSPD-MEF}, MESPD-MEF \cite{MESPD-MEF}, PESPD-MEF \cite{PESPD-MEF}, and ours.}
	\label{fig:multi_result02}
\end{figure*}	

\subsection{Model Verification}
The haze distribution perception model formulated in Section 2.2 integrates four principal features. In order to verify the validity of the model for haze distribution detection and the rationality of feature composition, this section conducts a systematic evaluation of its theoretical foundations.

The haze perception model integrates four core metrics ($BI$ (f1), $WC$ (f2), $CF$ (f3), and $V$ (f4)), each designed to capture a distinct aspect of the scattering phenomenon, from luminance elevation to contrast attenuation and statistical variance. The low Pearson Correlation Coefficients (PCC) and Mutual Information Coefficients (MIC) \cite{mic} between these features (Figure \ref{fig:hotmap}) confirm their complementarity in describing the physical properties of haze without redundant information. This principled feature selection ensures a comprehensive characterization of haze without unnecessary dimensionality.

Figure \ref{fig:score}(b) displays the normalized feature responses across five annotated image patches drawn from Figure \ref{fig:score}(a). By integrating these heterogeneous metrics, the model achieves improved objectivity in environmental assessment while retaining discriminative performance across varying haze densities. The resulting haze probability distribution is shown in Figure \ref{fig:haze}, where intensity corresponds to atmospheric particulate concentration. The spatial agreement between high-probability regions and observed haze accumulations in sample imagery confirms the model's accuracy in identifying zones of high haze concentration.

\subsection{Multi-Exposure Fusion Analysis}
The proposed framework is evaluated against five state-of-the-art methods: Mertens09 \cite{2009Exposure}, Wang19 \cite{Detail-Enhanced}, MSPD-MEF \cite{MSPD-MEF}, MESPD-MEF \cite{MESPD-MEF}, and PESPD-MEF \cite{PESPD-MEF}. The source codes for these algorithms can be accessed via the provided links in Ref.~\citenum{ZHANG2021} or directly from the respective authors. To ensure a fair comparison, all parameters of the compared methods are set to their default values.

\begin{table*}[t]
	\centering
	\small
	\caption{Quantitative results of multi-exposure fusion methods on four datasets.}
	\renewcommand\arraystretch{1.25} 
	\centering   
	\resizebox{0.9\textwidth}{!}{ 
		\begin{tabular}{p{4.04166650772095em}ccccccc}
			\toprule 
			\multicolumn{2}{c}{Metric} & \textbf{Mertens09} & \textbf{Wang19}& \textbf{MSPD-MEF} & \textbf{MESPD-MEF} & \textbf{PESPD-MEF} & \textbf{Ours} \\
			\midrule
			\multicolumn{2}{c}{\textit{AG} \cite{CUI2015199} $\uparrow$} & 3.2938  & 5.4696  & 5.0596  & 4.9045  & \underline{5.6689}  & \textbf{7.2503 } \\
			\multicolumn{2}{c}{\textit{VIF} \cite{VIF} $\uparrow$} & 0.8576 & 1.3408 & \underline{1.7327} & 1.2935 & 1.6943 & \textbf{1.7660 } \\
			\multicolumn{2}{c}{\textit{IE} \cite{IE} $\uparrow$} & 6.8199  & 7.1166  & 6.8395  & 7.3553  & \underline{7.4323}  & \textbf{7.4957 } \\
			\bottomrule 
			\multicolumn{8}{l}{ \footnotesize The best results are bolded, and the second-best are underlined. The symbol $\uparrow$ indicates that higher metric values are better.}
	\end{tabular}}%
	\label{tab:multi_fusion}%
\end{table*}

\subsubsection{Visualized Results}
As shown in Figure \ref{fig:multi_result01}, the dehazing results for \emph{Dataset1} are presented. Although all methods improve overall visibility, the limitations of conventional approaches in handling extreme physical conditions become apparent. Mertens09 \cite{2009Exposure} fails to preserve critical structural details. Both MSPD-MEF \cite{MSPD-MEF} and PESPD-MEF \cite{PESPD-MEF} introduce distortions due to overly aggressive contrast enhancement that violates the natural radiometric profile. MESPD-MEF \cite{MESPD-MEF} maintains global contrast but introduces noticeable artifacts (as highlighted by the yellow rectangle). Although the method proposed by Wang19 \cite{Detail-Enhanced} produces satisfactory results, our approach outperforms it in terms of both overall quality and visibility, as it integrates advanced image enhancement techniques with effective haze suppression.

\begin{figure*}[t]
	\centering	
	{\includegraphics[width=0.9\textwidth]{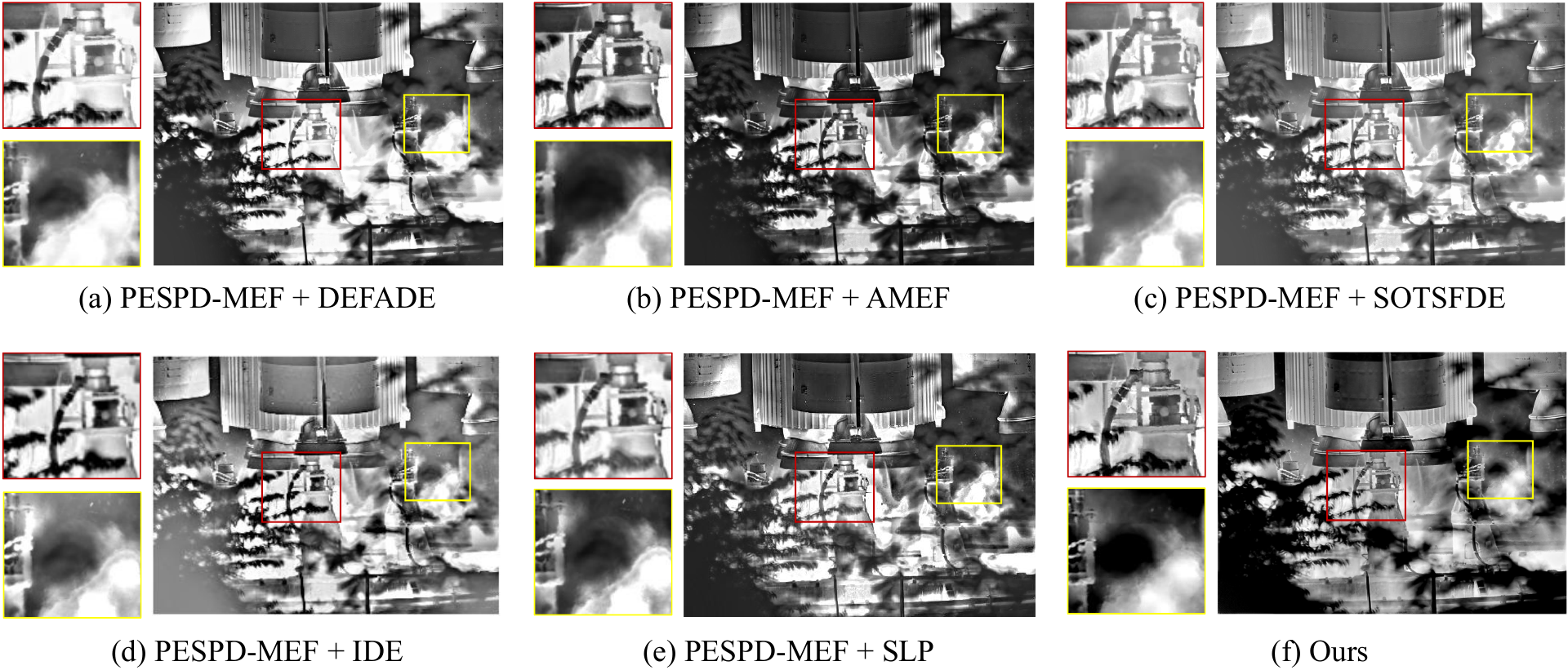}}
	\caption{Visual results of different dehazing methods applied to PESPD-MEF outputs, as shown in Figure \ref{fig:multi_result01}(f). From (a) to (f), results are reconstructed by the following methods: PESPD-MEF \cite{PESPD-MEF} + DEFADE \cite{FADE}, PESPD-MEF \cite{PESPD-MEF} + AMEF \cite{artificial2018}, PESPD-MEF \cite{PESPD-MEF} + SOTSFDE \cite{SOTSFDE}, PESPD-MEF \cite{PESPD-MEF} + IDE \cite{IDE}, PESPD-MEF \cite{PESPD-MEF} + SLP \cite{SLP} and ours.}
	\label{fig:dehaze_result01}
	\vspace{0.3cm}
\end{figure*}

In HDR combustion scenarios (Figure \ref{fig:multi_result02}), the challenge intensifies. Conventional methods fail to simultaneously retain the radiance of bright particles and suppress the scattering effects of haze. Mertens09 \cite{2009Exposure} leads to significant blurring, losing particulate information. MSPD-MEF \cite{MSPD-MEF} suffers from poor contrast, making it difficult to distinguish burning particles from the haze background. MESPD-MEF \cite{MESPD-MEF} and PESPD-MEF \cite{PESPD-MEF} incorrectly handle the flame's radiometric properties, causing the core to appear unnaturally gray. Wang19 \cite{Detail-Enhanced} retains contrast but fails to manage high-brightness regions, resulting in an unnatural appearance. In contrast, our method faithfully preserves the physical texture of the scene while achieving effective haze suppression, yielding images with enhanced detail and clarity that are more suitable for quantitative measurement.

\begin{figure*}[t]
	\centering
	\includegraphics[width=0.92\textwidth]{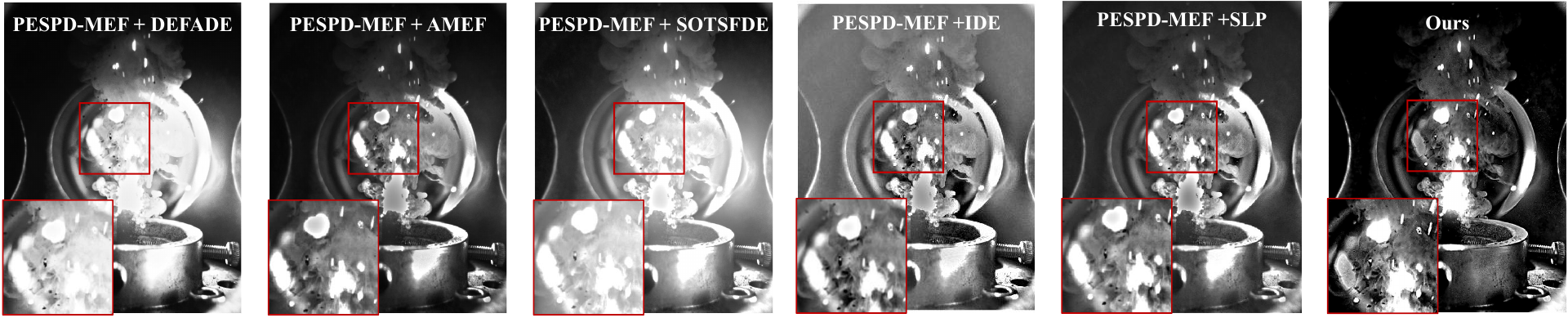}
	\caption{Visual results of different dehazing methods applied to PESPD-MEF outputs, as shown in Figure \ref{fig:multi_result02}(f). From (a) to (f), results are reconstructed by the following methods: PESPD-MEF \cite{PESPD-MEF} + DEFADE \cite{FADE}, PESPD-MEF \cite{PESPD-MEF} + AMEF \cite{artificial2018}, PESPD-MEF \cite{PESPD-MEF} + SOTSFDE \cite{SOTSFDE}, PESPD-MEF \cite{PESPD-MEF} + IDE \cite{IDE}, PESPD-MEF \cite{PESPD-MEF} + SLP \cite{SLP} and ours.}
	\label{fig:dehaze_result02}
\end{figure*}

\subsubsection{Objective Evaluation Results}
The proposed method is evaluated using three standard image fusion metrics: Average Gradient (AG) \cite{CUI2015199}, Visual Information Fidelity (VIF), and Information Entropy (IE) \cite{IE}. Results in Table \ref{tab:multi_fusion} show that our method consistently outperforms others across all datasets (including haze-free \emph{Dataset3}), with best scores bolded and second-best underlined.
Our method achieves the highest AG, indicating sharper edges and better-preserved detail critical for feature extraction. The best VIF confirms superior visual quality and perceptual alignment, while the highest IE reflects maximal retention of source information for downstream analysis. Overall, the framework excels in haze suppression while maintaining structural and photometric fidelity.

\subsection{Haze Suppression Experiments}
Typical multi-exposure fusion methods do not integrate haze suppression capabilities. To address this limitation, the current method is combined with several state-of-the-art dehazing techniques for both qualitative and quantitative comparisons. PESPD-MEF, renowned for its strong overall performance, is selected for comparison with five recent dehazing methods: DEFADE \cite{FADE}, AMEF \cite{artificial2018}, SOTSFDE \cite{SOTSFDE}, IDE \cite{IDE}, and SLP \cite{SLP}.

\subsubsection{Visualized Results}
Figure \ref{fig:dehaze_result01} compares several dehazing methods applied to the PESPD-MEF result from Figure \ref{fig:multi_result01}(f). While most methods enhance details originally obscured by haze, AMEF \cite{artificial2018} and SLP \cite{SLP} erroneously suppress non-haze regions like the rocket body, violating scene physics. DEFADE \cite{FADE} reduces haze effectively but leaves residual haze in certain areas, as shown in the red rectangle in Figure \ref{fig:dehaze_result01}(a). SOTSFDE \cite{SOTSFDE} recovers the overall scene structure but falls short in reconstructing fine details. IDE \cite{IDE} improves visibility yet fails to deliver well-balanced contrast, particularly in the region marked by the red rectangle in Figure \ref{fig:dehaze_result01}(d). Critically, the software-simulated multi-exposure in AMEF cannot replicate the authentic nonlinear radiometric responses captured by our physical SVE hardware, leading to its failure under extreme combustion conditions.

Figure \ref{fig:dehaze_result02} further demonstrates enhanced image quality in combustion scenes relative to Figure \ref{fig:multi_result02}(f). Both PESPD-MEF \cite{PESPD-MEF} and SOTSFDE \cite{SOTSFDE} exhibit limited performance under intense haze from combustion. Although AMEF \cite{artificial2018} competently removes haze, its software-based exposure synthesis cannot faithfully represent the complex radiometric-scattering interplay in high-intensity flames, resulting in a loss of particulate textural details (see the red rectangle in Figure \ref{fig:dehaze_result02}(b)). IDE \cite{IDE} suffers from inconsistent contrast, while SLP \cite{SLP}, though effective at haze removal, misclassifies the flame as haze and introduces artifacts. In contrast, our approach, rooted in real multi-exposure data, accurately detects and suppresses haze while preserving critical combustion details.

\subsubsection{No-Reference Image Quality Assessment}
To quantitatively evaluate dehazing performance, we employ two widely used no-reference metrics: the Haze-Aware Density Evaluator (FADE) \cite{FADE} and the No-Reference Image Quality Metric for Contrast Distortion (NIQMC) \cite{NIQMC}. The results for various methods are summarized in Table \ref{tab:dehaze-1}, with the best and second-best scores highlighted in bold and underlined, respectively.

Our method consistently ranks first or second across all metrics, demonstrating robust performance in varied hazy conditions. While AMEF \cite{artificial2018} shows competitive FADE scores in some cases, it struggles to preserve details in challenging scenarios. Although SLP \cite{SLP} achieves a marginally better FADE score, our method provides a better balance between haze removal and contrast preservation, as reflected in its superior NIQMC score---critical for both visual and automated analysis. This slight trade-off in FADE stems from our joint brightness adjustment and dehazing process, which better maintains perceptual quality and detail. Overall, the results confirm the robustness and effectiveness of our framework for haze suppression and image enhancement.

\begin{table}[tp]
	\centering
	\caption{Quantitative dehazing results of different methods.}
	\renewcommand\arraystretch{1.5} 
	\resizebox{0.65\textwidth}{!}{
		\begin{threeparttable}  
			\begin{tabular}{ccccc}
				\toprule
				Method & \multicolumn{2}{c}{\textit{FADE} \citep{FADE} $\downarrow$} & \multicolumn{2}{c}{\textit{NIQMC} \citep{NIQMC} $\uparrow$} \\
				\midrule
				\textbf{PESPD-MEF} & \multicolumn{2}{c}{1.2462} & \multicolumn{2}{c}{5.0583 } \\
				\textbf{PESPD-MEF} + \textbf{DEFADE} & \multicolumn{2}{c}{1.0925 } & \multicolumn{2}{c}{5.1514 } \\
				\textbf{PESPD-MEF} + \textbf{AMEF} & \multicolumn{2}{c}{0.7733 } & \multicolumn{2}{c}{4.9853 } \\
				\textbf{PESPD-MEF} + \textbf{SOTSFDE} & \multicolumn{2}{c}{1.2483 } & \multicolumn{2}{c}{5.0433 } \\
				\textbf{PESPD-MEF} + \textbf{IDE} & \multicolumn{2}{c}{0.7674 } & \multicolumn{2}{c}{5.4560 } \\
				\textbf{PESPD-MEF} + \textbf{SLP} & \multicolumn{2}{c}{\textbf{0.7494} } & \multicolumn{2}{c}{\underline{5.4811}} \\
				\textbf{Our method} & \multicolumn{2}{c}{\underline{0.7561} } & \multicolumn{2}{c}{\textbf{5.6331}} \\
				\bottomrule
			\end{tabular}%
			\begin{tablenotes}            
				\item The best results are bolded, and the second-best are underlined. The symbol $\uparrow$ indicates that higher metric values are better, while $\downarrow$ indicates that lower values are better.
			\end{tablenotes}  
		\end{threeparttable}
		\label{tab:dehaze-1}}
\end{table}

\subsubsection{Full-Reference Image Quality Assessment}
\begin{figure}[H]
	\centering
	\includegraphics[width=0.5\textwidth]{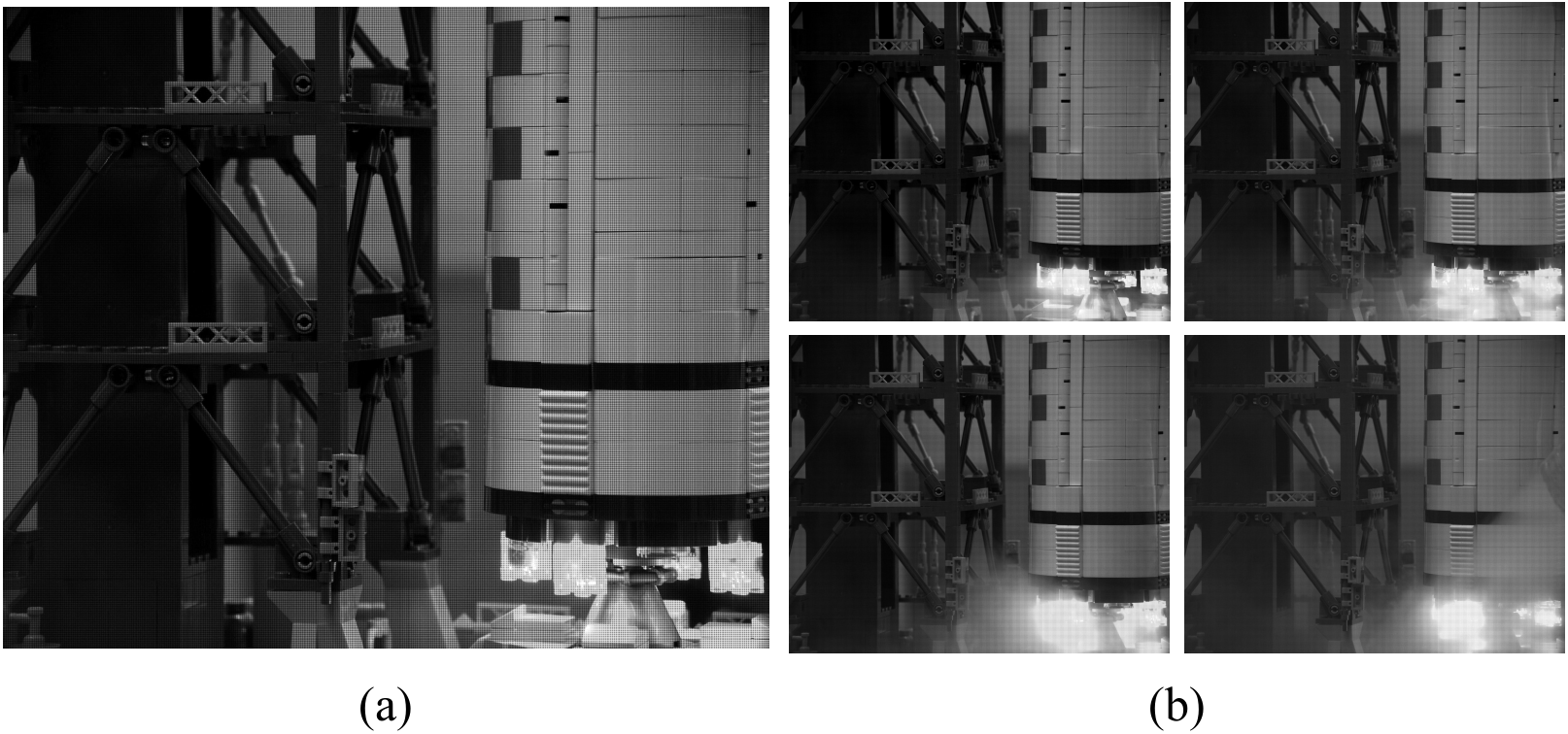}
	\caption{Some test images from \emph{Dataset3}: (a) Original image; (b) Images with varying degrees of added haze.}
	\label{fig:simulate}
	\vspace{0.3cm}
	\centering
	\includegraphics[width=0.5\textwidth]{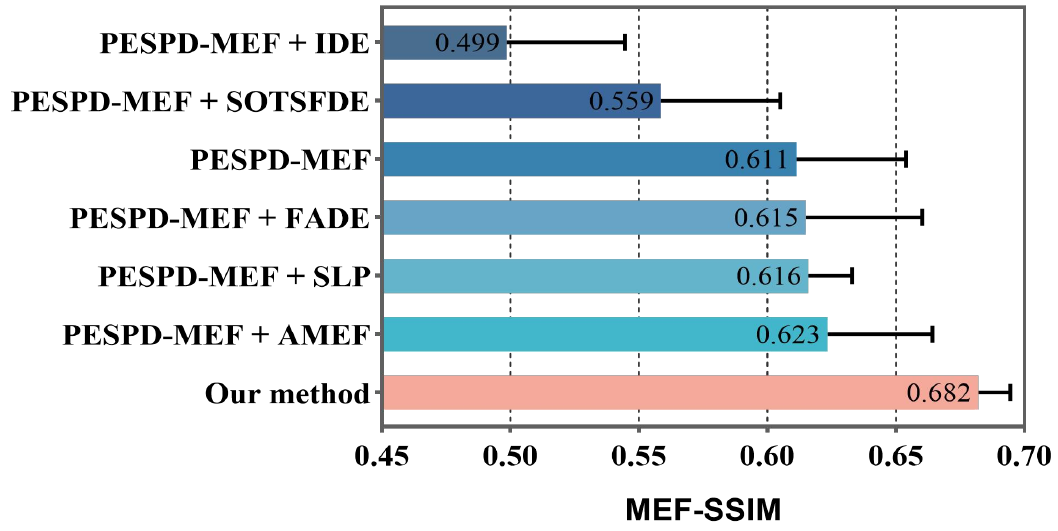}
	\caption{Average MEF-SSIM \cite{mef-ssim} metrics of different methods on the simulation dataset (\emph{Dataset3}). From top to bottom, the method scores gradually became better.}
	\label{fig:mef-ssim}
\end{figure}

For full-reference evaluation, it is assumed that hazy images and their corresponding haze-free versions are available. The processed image is compared to the original to assess quality improvements. For this, we simulate a rocket launch scenario (\emph{Dataset3}) with various haze types. Figure \ref{fig:simulate} shows a sample image from the dataset, along with hazy images generated under four different conditions.
The MEF-based structural similarity index measure (MEF-SSIM) \cite{mef-ssim} quantifies similarity between haze-free and dehazed images. As shown in Figure \ref{fig:mef-ssim}, our method outperforms others in MEF-SSIM, confirming its effectiveness. Furthermore, our method demonstrates reduced variability across different haze types, highlighting its robust performance. It is noteworthy that some hybrid methods exhibit inferior performance compared to the multi-exposure fusion technique alone, likely due to excessive suppression in non-hazy regions and the introduction of halo artifacts.

\subsection{Ablation Experiments}
The proposed framework integrates two core components: sub-regional adaptive enhancement (Module A) and adaptive sub-region weight estimation (Module B). To systematically evaluate the contributions of each module, ablation studies were conducted by selectively deactivating components and analyzing their impact on performance.

\begin{figure*}[t]
	\centering	
	{\includegraphics[width=0.8\textwidth]{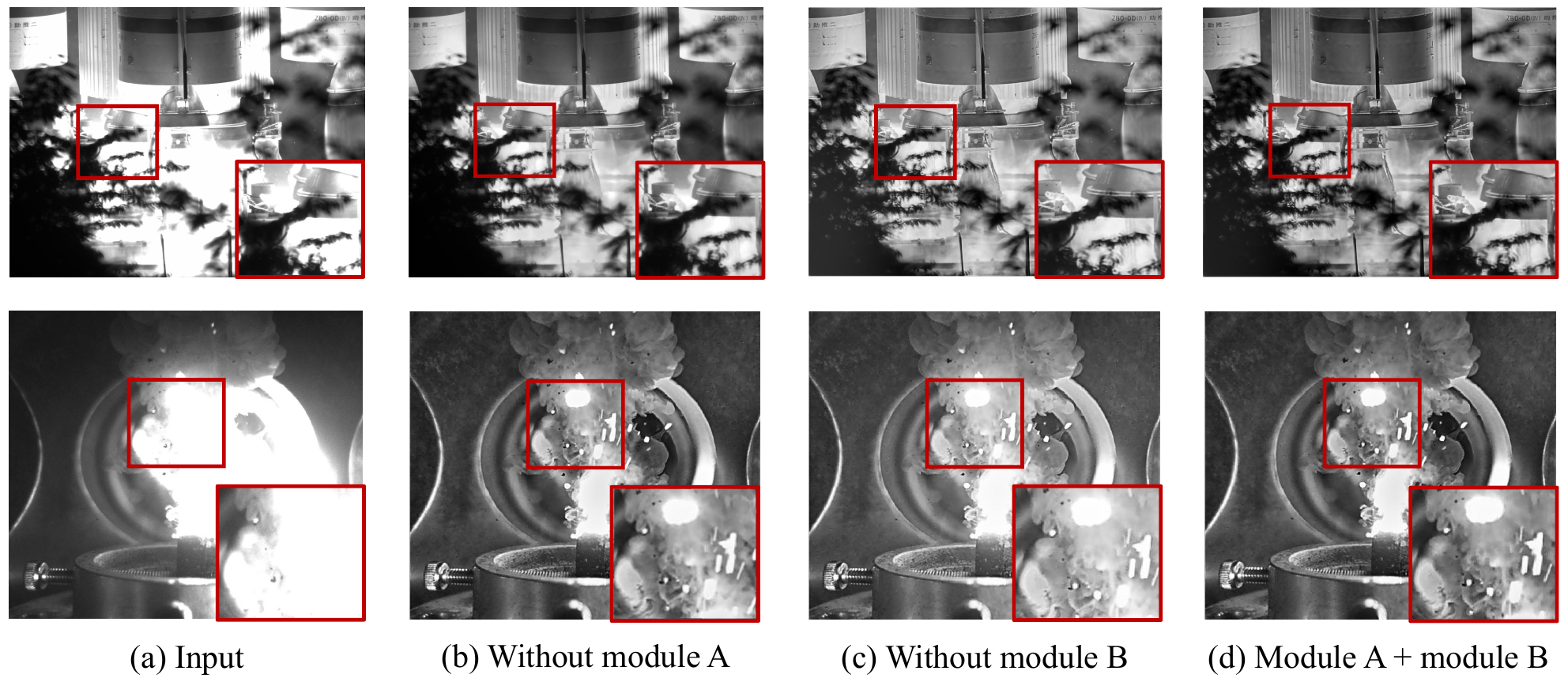}}
	\caption{Dehazing ablation study. (a) Input images. (b) Results without the adaptive sub-region enhancement module (Module A). (c) Results without the adaptive sub-region weight calculation module (Module B). (d) Results using the proposed framework.}
	\label{fig:ablation}
	\vspace{0.2cm}
	\centering
	\includegraphics[width=0.72\textwidth]{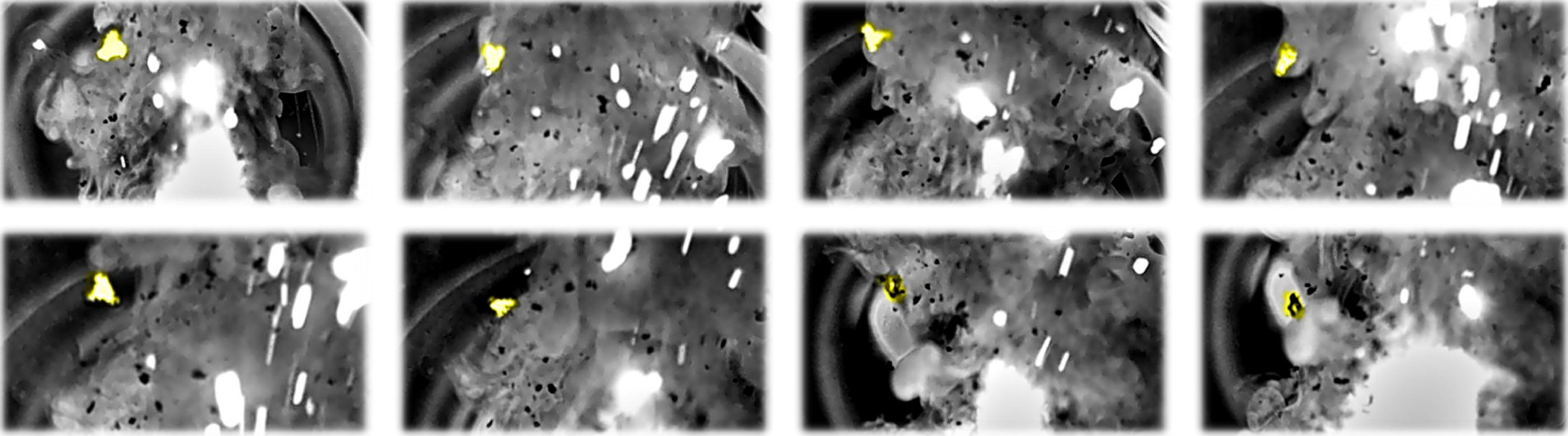}
	\caption{Image sequence depicting particle detection using recovered images from our proposed method.}
	\label{fig:application}
\end{figure*}

As shown in Figure \ref{fig:ablation}(b), replacing Module A with a global enhancement strategy impairs the system’s capacity for localized photometric compensation, leading to insufficient highlight suppression and detail recovery in regions affected by strong scattering. In Figure \ref{fig:ablation}(c), disabling Module B---which accounts for haze distribution in weight calculation---results in poor local contrast and ineffective haze suppression, due to non-adaptive fusion across exposures. Conversely, the integrated framework in Figure \ref{fig:ablation}(d) achieves comprehensive atmospheric correction through: (1) photometrically consistent global illumination adjustment, (2) edge-preserving local contrast enhancement, and (3) region-aware suppression of scattering effects.

These results affirm that both modules are essential for optimal performance, demonstrating the necessity of combining localized enhancement with haze-aware weight estimation for effective dehazing.

\subsection{Application in Combustion Diagnostics}
To further demonstrate the value of our framework beyond qualitative enhancement, we apply it to quantitative combustion diagnostics—a core area of experimental mechanics. As quantitatively validated in Figure \ref{fig:multi_result02}, \ref{fig:dehaze_result02} and \ref{fig:application}, the enhanced dynamic range and clarity enable the detection and tracking of particle combustion and extinction states---key parameters for fuel performance research \cite{review,SUN2025135969,photofragmentation}.

Furthermore, the integration of multi-view observations with relative pose estimation algorithms can potentially enable 3D reconstruction of particle trajectories under haze-affected conditions \cite{guanpami25}. This advancement paves the way for multi-modal combustion diagnostics by combining optical imaging with computational fluid dynamics simulations, highlighting the framework's adaptability to interdisciplinary challenges in propulsion engineering and environmental monitoring.

\section{Conclusion}
Optical measurement of critical mechanical parameters such as plume flow fields, shock wave structures, and nozzle oscillations during rocket launch is severely compromised by extreme illumination conditions and dense combustion haze. To address these challenges, we developed a novel dehazing framework that integrates multi-exposure fusion with physics-aware haze suppression through a custom SVE camera. This co-designed hardware-software system captures multi-exposure data in a single shot, enabling adaptive haze removal and high-fidelity image reconstruction. The proposed method effectively recovers visual information essential for mechanical analysis, significantly enhancing detail preservation and contrast under extreme radiative conditions. Experimental results demonstrate its capability to accurately extract key parameters including particle velocity and structural vibrations. This work provides a reliable imaging foundation for quantitative mechanical analysis in rocket launch applications.

\section{Limitations and Future Work}
While our framework demonstrates strong performance, we identify several avenues for future research. Future efforts will focus on enhancing real-time processing capabilities for onboard deployment via hardware acceleration (e.g., FPGA/GPU) and exploring multi-modal sensor integration. Further avenues include developing next-generation SVE sensors for greater dynamic ranges; adapting the framework to other harsh environments, such as underwater or dust-filled scenes, would require re-optimizing the SVE design and haze perception model. Finally, robust, automated in-situ calibration techniques will be critical for ensuring field reliability.

\subsection*{Acknowledgement}
This work was supported by the Hunan Provincial Natural Science Foundation for Excellent Young Scholars (Grant No. 2023JJ20045), the Foundation of National Key Laboratory of Human Factors Engineering (Grant No. GJSD22006), and the National Natural Science Foundation of China (Grant No. 12372189).

\subsection*{Disclosures}
The authors declare no conflicts of interest.


\begin{thebibliography}{99}
	\providecommand{\natexlab}[1]{#1}
	\providecommand{\url}[1]{\texttt{#1}}
	\expandafter\ifx\csname urlstyle\endcsname\relax
	\providecommand{\doi}[1]{doi: #1}\else
	\providecommand{\doi}{doi: \begingroup \urlstyle{rm}\Url}\fi
	
	\bibitem[Karr et~al.(2016)Karr, Chalmers, and Debattista]{2014High}
	B.~Karr, A.~Chalmers, and K.~Debattista.
	\newblock Chapter 20 - High Dynamic Range Digital Imaging of Spacecraft.
	\newblock In \emph{High Dynamic Range Video}, pages 519--547. Academic Press, 2016.
	\newblock ISBN 978-0-08-100412-8.
	
	\bibitem[Zhao and Ladommatos(1998)]{ZHAO1998221}
	H.~Zhao and N.~Ladommatos.
	\newblock Optical diagnostics for soot and temperature measurement in diesel engines.
	\newblock \emph{Progress in Energy and Combustion Science}, 24\penalty0 (3):\penalty0 221--255, 1998.
	\newblock ISSN 0360-1285.
	\newblock \doi{10.1016/S0360-1285(97)00033-6}.
	
	\bibitem[Wang et~al.(2021)Wang, Zhang, and Zhu]{Removal2021}
	Xin Wang, Xin Zhang, Hangcheng Zhu, et~al.
	\newblock An Effective Algorithm for Single Image Fog Removal.
	\newblock \emph{Mobile Networks and Applications}, 26\penalty0 (3):\penalty0 1250--1258, Jun 2021.
	\newblock ISSN 1383-469X.
	
	\bibitem[Wei et~al.(2021)Wei, Wu, and Li]{Non-Homogeneous}
	Haoran Wei, Qingbo Wu, Hui Li, et~al.
	\newblock Non-Homogeneous Haze Removal via Artificial Scene Prior and Bidimensional Graph Reasoning.
	\newblock \emph{IEEE Transactions on Image Processing}, 30:\penalty0 9136--9149, 2021.
	
	\bibitem[Ding et~al.(2024)Ding, Zhang, and Xu]{U2D2Net}
	Bosheng Ding, Ruiheng Zhang, Lixin Xu, et~al.
	\newblock U2D2Net: Unsupervised Unified Image Dehazing and Denoising Network for Single Hazy Image Enhancement.
	\newblock \emph{IEEE Transactions on Multimedia}, 26:\penalty0 202--217, 2024.
	
	\bibitem[Gao et~al.(2023)Gao, Guan, and Zhang]{Gao23}
	Liuzheng Gao, Banglei Guan, Li Zhang, et~al.
	\newblock Two large-exposure-ratio image fusion by improved morphological segmentation.
	\newblock \emph{Applied optics}, 62\penalty0 (29):\penalty0 7713--7720, 2023.
	
	\bibitem[Galdran(2018)]{artificial2018}
	A.~Galdran.
	\newblock Image dehazing by artificial multiple-exposure image fusion.
	\newblock \emph{Signal Processing}, 149:\penalty0 135--147, 2018.
	\newblock ISSN 0165-1684.
	
	\bibitem[He et~al.(2011)He, Sun, and Tang]{DCP}
	Kaiming He, Jian Sun, and Xiaoou Tang.
	\newblock Single Image Haze Removal Using Dark Channel Prior.
	\newblock \emph{IEEE Transactions on Pattern Analysis and Machine Intelligence}, 33\penalty0 (12):\penalty0 2341--2353, 2011.
	
	\bibitem[Ling et~al.(2023)Ling, Chen, and Tan]{SLP}
	Pengyang Ling, Huaian Chen, Xiao Tan, et~al.
	\newblock Single Image Dehazing Using Saturation Line Prior.
	\newblock \emph{IEEE Transactions on Image Processing}, 32:\penalty0 3238--3253, 2023.
	
	\bibitem[Zhou et~al.(2022)Zhou, Liu, and Duan]{PSD}
	Shuyi Zhou, Xiaoyan Liu, Jiaxu Duan, et~al.
	\newblock A Novel Model-Based Defogging Method for Particle Images With Different Fog Distributions.
	\newblock \emph{IEEE Transactions on Instrumentation and Measurement}, 71:\penalty0 1--19, 2022.
	
	\bibitem[Jackson et~al.(2017)Jackson, Ariyo, and Acheampong]{2017ICIVC}
	Jehoiada Jackson, Oluwasanmi Ariyo, Kingsley Acheampong, et~al.
	\newblock Hybrid single image dehazing with bright channel and dark channel priors.
	\newblock In \emph{International Conference on Image, Vision and Computing (ICIVC)}, pages 381--385, 2017.
	
	\bibitem[Kinoshita and Kiya(2019)]{2019Scene}
	Yuma Kinoshita and Hitoshi Kiya.
	\newblock Scene Segmentation-Based Luminance Adjustment for Multi-Exposure Image Fusion.
	\newblock \emph{IEEE Transactions on Image Processing}, 28\penalty0 (8):\penalty0 4101--4116, 2019.
	
	\bibitem[Cantor(1978)]{Optic}
	A.~Cantor.
	\newblock Optics of the atmosphere-Scattering by molecules and particles.
	\newblock \emph{IEEE Journal of Quantum Electronics}, 14\penalty0 (9):\penalty0 698--699, 1978.
	
	\bibitem[Lee et~al.(2018)Lee, Park, and Cho]{ICIP}
	Sang-hoon Lee, Jae Sung Park, and Nam Ik Cho.
	\newblock A Multi-Exposure Image Fusion Based on the Adaptive Weights Reflecting the Relative Pixel Intensity and Global Gradient.
	\newblock In \emph{IEEE International Conference on Image Processing (ICIP)}, pages 1737--1741, 2018.
	
	\bibitem[Tao et~al.(2024)Tao, Li, and Guan]{Simultaneous}
	Jing Tao, You Li, Banglei Guan, et~al.
	\newblock Simultaneous Enhancement and Noise Suppression Under Complex Illumination Conditions.
	\newblock \emph{IEEE Transactions on Instrumentation and Measurement}, 73:\penalty0 1--11, 2024.
	
	\bibitem[Berman et~al.(2016)Berman, Treibitz, and Avidan]{Non-local}
	Dana Berman, Tali Treibitz, and Shai Avidan.
	\newblock Non-local Image Dehazing.
	\newblock In \emph{IEEE Conference on Computer Vision and Pattern Recognition (CVPR)}, pages 1674--1682, 2016.
	
	\bibitem[Ma et~al.(2018)Ma, Duanmu, and Yeganeh]{2018Multi-Exposure}
	Kede Ma, Zhengfang Duanmu, Hojatollah Yeganeh, et~al.
	\newblock Multi-Exposure Image Fusion by Optimizing A Structural Similarity Index.
	\newblock \emph{IEEE Transactions on Computational Imaging}, 4\penalty0 (1):\penalty0 60--72, 2018.
	
	\bibitem[Wang et~al.(2022)Wang, Wang, and Liu]{Variational}
	Wenhui Wang, Anna Wang, and Chen Liu.
	\newblock Variational Single Nighttime Image Haze Removal With a Gray Haze-Line Prior.
	\newblock \emph{IEEE Transactions on Image Processing}, 31:\penalty0 1349--1363, 2022.
	
	\bibitem[Saponara(2016)]{2016Real}
	Sergio Saponara.
	\newblock Real-time Multi-camera Video Acquisition and Processing Platform for ADAS.
	\newblock In \emph{Conference on real-time image and video processing}, volume 9897, 2016.
	
	\bibitem[Tescher et~al.(2016)Tescher, Shopovska, and Jovanov]{2016HDR}
	Andrew~G. Tescher, Ivana Shopovska, Ljubomir Jovanov, et~al.
	\newblock {HDR} video synthesis for vision systems in dynamic scenes.
	\newblock \emph{Applications of Digital Image Processing}, 9971:\penalty0 99710C, 2016.
	
	\bibitem[Yu et~al.(2024)Yu, Bu, and Zhang]{Spikes}
	Zhaofei Yu, Tong Bu, Yijun Zhang, et~al.
	\newblock Robust Decoding of Rich Dynamical Visual Scenes With Retinal Spikes.
	\newblock \emph{IEEE Transactions on Neural Networks and Learning Systems}, pages 1--14, 2024.
	
	\bibitem[Wang et~al.(2013)Wang, Zhuo, and Tao]{Wang2013}
	Yinting Wang, Shaojie Zhuo, Dapeng Tao, et~al.
	\newblock Automatic local exposure correction using bright channel prior for under-exposed images.
	\newblock \emph{Signal Processing}, 93\penalty0 (11):\penalty0 3227--3238, 2013.
	\newblock ISSN 0165-1684.
	
	\bibitem[Kou et~al.(2015)Kou, Chen, and Wen]{WGIF}
	Fei Kou, Weihai Chen, Changyun Wen, et~al.
	\newblock Gradient Domain Guided Image Filtering.
	\newblock \emph{IEEE Transactions on Image Processing}, 24\penalty0 (11):\penalty0 4528--4539, 2015.
	
	\bibitem[Mertens et~al.(2009)Mertens, Kautz, and Reeth]{2009Exposure}
	Tom Mertens, Jan Kautz, and Frank~Van Reeth.
	\newblock Exposure Fusion: A Simple and Practical Alternative to High Dynamic Range Photography.
	\newblock \emph{Computer Graphics Forum}, 28\penalty0 (1):\penalty0 161--171, 2009.
	
	\bibitem[Wang et~al.(2020)Wang, Chen, and Wu]{Detail-Enhanced}
	Qiantong Wang, Weihai Chen, Xingming Wu, et~al.
	\newblock Detail-Enhanced Multi-Scale Exposure Fusion in YUV Color Space.
	\newblock \emph{IEEE Transactions on Circuits and Systems for Video Technology}, 30\penalty0 (8):\penalty0 2418--2429, 2020.
	
	\bibitem[Li et~al.(2020)Li, Ma, and Yong]{MSPD-MEF}
	Hui Li, Kede Ma, Hongwei Yong, et~al.
	\newblock Fast Multi-Scale Structural Patch Decomposition for Multi-Exposure Image Fusion.
	\newblock \emph{IEEE Transactions on Image Processing}, 29:\penalty0 5805--5816, 2020.
	
	\bibitem[Li et~al.(2021)Li, Chan, and Qi]{MESPD-MEF}
	Hui Li, Tsz~Nam Chan, Xianbiao Qi, et~al.
	\newblock Detail-Preserving Multi-Exposure Fusion With Edge-Preserving Structural Patch Decomposition.
	\newblock \emph{IEEE Transactions on Circuits and Systems for Video Technology}, 31\penalty0 (11):\penalty0 4293--4304, 2021.
	
	\bibitem[Zhang et~al.(2023)Zhang, Luo, and Huang]{PESPD-MEF}
	Junchao Zhang, Yidong Luo, Junbin Huang, et~al.
	\newblock Multi-exposure image fusion via perception enhanced structural patch decomposition.
	\newblock \emph{Information Fusion}, 99:\penalty0 101895, 2023.
	\newblock ISSN 1566-2535.
	
	\bibitem[Zhang(2021)]{ZHANG2021}
	Xingchen Zhang.
	\newblock Benchmarking and comparing multi-exposure image fusion algorithms.
	\newblock \emph{Information Fusion}, 74:\penalty0 111--131, 2021.
	\newblock ISSN 1566-2535.
	
	\bibitem[Cui et~al.(2015)Cui, Feng, and Xu]{CUI2015199}
	Guangmang Cui, Huajun Feng, Zhihai Xu, et~al.
	\newblock Detail preserved fusion of visible and infrared images using regional saliency extraction and multi-scale image decomposition.
	\newblock \emph{Optics Communications}, 341:\penalty0 199--209, 2015.
	\newblock ISSN 0030-4018.
	
	\bibitem[Sheikh and Bovik(2006)]{VIF}
	H.R. Sheikh and A.C. Bovik.
	\newblock Image information and visual quality.
	\newblock \emph{IEEE Transactions on Image Processing}, 15\penalty0 (2):\penalty0 430--444, 2006.
	
	\bibitem[Roberts et~al.(2008)Roberts, van Aardt, and Ahmed]{IE}
	Wesley Roberts, Jan van Aardt, and Fethi Ahmed.
	\newblock Assessment of image fusion procedures using entropy, image quality, and multispectral classification.
	\newblock \emph{Journal of Applied Remote Sensing}, 2:\penalty0 1--28, May 2008.
	\newblock \doi{10.1117/1.2945910}.
	
	\bibitem[Choi et~al.(2015)Choi, You, and Bovik]{FADE}
	Lark~Kwon Choi, Jaehee You, and Alan~Conrad Bovik.
	\newblock Referenceless Prediction of Perceptual Fog Density and Perceptual Image Defogging.
	\newblock \emph{IEEE Transactions on Image Processing}, 24\penalty0 (11):\penalty0 3888--3901, 2015.
	
	\bibitem[Ling et~al.(2018)Ling, Gong, and Fan]{SOTSFDE}
	Zhigang Ling, Jianwei Gong, Guoliang Fan, et~al.
	\newblock Optimal Transmission Estimation via Fog Density Perception for Efficient Single Image Defogging.
	\newblock \emph{IEEE Transactions on Multimedia}, 20\penalty0 (7):\penalty0 1699--1711, 2018.
	
	\bibitem[Ju et~al.(2021)Ju, Ding, and Ren]{IDE}
	Mingye Ju, Can Ding, Wenqi Ren, et~al.
	\newblock {IDE}: Image Dehazing and Exposure Using an Enhanced Atmospheric Scattering Model.
	\newblock \emph{IEEE Transactions on Image Processing}, 30:\penalty0 2180--2192, 2021.
	
	\bibitem[Gu et~al.(2017)Gu, Lin, and Zhai]{NIQMC}
	Ke Gu, Weisi Lin, Guangtao Zhai, et~al.
	\newblock No-Reference Quality Metric of Contrast-Distorted Images Based on Information Maximization.
	\newblock \emph{IEEE Transactions on Cybernetics}, 47\penalty0 (12):\penalty0 4559--4565, 2017.
	
	\bibitem[Ma et~al.(2015)Ma, Zeng, and Wang]{mef-ssim}
	Kede Ma, Kai Zeng, and Zhou Wang.
	\newblock Perceptual Quality Assessment for Multi-Exposure Image Fusion.
	\newblock \emph{IEEE Transactions on Image Processing}, 24\penalty0 (11):\penalty0 3345--3356, 2015.
	
	\bibitem[Ma et~al.(2017)Ma, Li, and Yong]{SPD-MEF}
	Kede Ma, Hui Li, Hongwei Yong, et~al.
	\newblock Robust Multi-Exposure Image Fusion: A Structural Patch Decomposition Approach.
	\newblock \emph{IEEE Transactions on Image Processing}, 26\penalty0 (5):\penalty0 2519--2532, 2017.
	
	\bibitem[Han et~al.(2022)Han, Li, and Guo]{DPE-MEF}
	Dong Han, Liang Li, Xiaojie Guo, et~al.
	\newblock Multi-exposure image fusion via deep perceptual enhancement.
	\newblock \emph{Information Fusion}, 79:\penalty0 248--262, 2022.
	\newblock ISSN 1566-2535.
	
	\bibitem[Li et~al.(2023)Li, Liu, and Zhou]{Detail-Refinement}
	Jiawei Li, Jinyuan Liu, Shihua Zhou, et~al.
	\newblock Learning a Coordinated Network for Detail-Refinement Multiexposure Image Fusion.
	\newblock \emph{IEEE Transactions on Circuits and Systems for Video Technology}, 33\penalty0 (2):\penalty0 713--727, 2023.
	
	\bibitem[Yang et~al.(2022)Yang, Wang, and Liu]{2017DeepFuse}
	Yang Yang, Chaoyue Wang, Risheng Liu, et~al.
	\newblock Self-augmented Unpaired Image Dehazing via Density and Depth Decomposition.
	\newblock In \emph{IEEE/CVF Conference on Computer Vision and Pattern Recognition (CVPR)}, pages 2027--2036, 2022.
	
	\bibitem[Chu et~al.(2022)Chu, Chen, and Fu]{atmos}
	Ying Chu, Fan Chen, Hong Fu, et~al.
	\newblock Haze Level Evaluation Using Dark and Bright Channel Prior Information.
	\newblock \emph{Atmosphere}, 13\penalty0 (5):\penalty0 683, 2022.
	\newblock ISSN 2073-4433.
	
	\bibitem[Gao et~al.(2024)Gao, Luo, and Wang]{Multiscale}
	Li Gao, DeLin Luo, and Song Wang.
	\newblock Multiscale feature learning and attention mechanism for infrared and visible image fusion.
	\newblock \emph{Science China(Technological Sciences)}, pages 408--422, 2024.
	
	\bibitem[Liu et~al.(2023)Liu, Guo, and Lu]{Network-Enabled}
	Ryan~Wen Liu, Yu Guo, Yuxu Lu, et~al.
	\newblock Deep Network-Enabled Haze Visibility Enhancement for Visual IoT-Driven Intelligent Transportation Systems.
	\newblock \emph{IEEE Transactions on Industrial Informatics}, 19\penalty0 (2):\penalty0 1581--1591, 2023.
	
	\bibitem[Qu et~al.(2023)Qu, Yin, and Liu]{AIM-MEF}
	Linhao Qu, Siqi Yin, Shaolei Liu, et~al.
	\newblock AIM-MEF: Multi-exposure image fusion based on adaptive information mining in both spatial and frequency domains.
	\newblock \emph{Expert Systems with Applications}, 223:\penalty0 119909, 2023.
	\newblock ISSN 0957-4174.
	
	\bibitem[Liu et~al.(2024)Liu, Liao, and Jiang]{Multi-exposure2024}
	Yun Liu, Guanglong Liao, Gangyi Jiang, et~al.
	\newblock Multi-exposure fused light field image quality assessment for dynamic scenes: Benchmark dataset and objective metric.
	\newblock \emph{Expert Systems with Applications}, 256:\penalty0 124881, 2024.
	\newblock ISSN 0957-4174.
	
	\bibitem[Nayar and Mitsunaga(2000)]{varying}
	S.K. Nayar and T.~Mitsunaga.
	\newblock High dynamic range imaging: spatially varying pixel exposures.
	\newblock In \emph{Proceedings IEEE Conference on Computer Vision and Pattern Recognition. CVPR 2000}, volume~1, pages 472--479, 2000.
	\newblock \doi{10.1109/CVPR.2000.855857}.
	
	\bibitem[Nayar and Mitsunaga(2000)]{SVE}
	S.K. Nayar and T.~Mitsunaga.
	\newblock High dynamic range imaging: spatially varying pixel exposures.
	\newblock In \emph{Proceedings IEEE Conference on Computer Vision and Pattern Recognition. CVPR 2000}, volume~1, pages 472--479, 2000.
	\newblock \doi{10.1109/CVPR.2000.855857}.
	
	\bibitem[Qu et~al.(2024)Qu, Chi, and Chan]{Spatially}
	Xiangyu Qu, Yiheng Chi, and Stanley~H. Chan.
	\newblock Spatially Varying Exposure With 2-by-2 Multiplexing: Optimality and Universality.
	\newblock \emph{IEEE Transactions on Computational Imaging}, 10:\penalty0 261--276, 2024.
	\newblock \doi{10.1109/TCI.2024.3354426}.
	
	\bibitem[Ye et~al.(2022)Ye, Gao, and Guan]{IROS2022}
	Keyang Ye, Liuzheng Gao, and Banglei Guan.
	\newblock Visual Odometry in HDR Environments by Using Spatially Varying Exposure Camera.
	\newblock In \emph{2022 IEEE/RSJ International Conference on Intelligent Robots and Systems (IROS)}, pages 5995--6000, 2022.
	\newblock \doi{10.1109/IROS47612.2022.9981538}.
	
	\bibitem[Nurit et~al.(2021)Nurit, {Le Goïc}, and Lewis]{NURIT2021103500}
	Marvin Nurit, Gaëtan {Le Goïc}, David Lewis, et~al.
	\newblock HD-RTI: An adaptive multi-light imaging approach for the quality assessment of manufactured surfaces.
	\newblock \emph{Computers in Industry}, 132:\penalty0 103500, 2021.
	\newblock \doi{10.1016/j.compind.2021.103500}.
	
	\bibitem[Reshef et~al.(2011)Reshef, Reshef, and Finucane]{mic}
	David~N. Reshef, Yakir~A. Reshef, Hilary~K. Finucane, et~al.
	\newblock Detecting Novel Associations in Large Data Sets.
	\newblock \emph{Science}, 334\penalty0 (6062):\penalty0 1518--1524, 2011.
	\newblock \doi{10.1126/science.1205438}.
	
	\bibitem[Mittal et~al.(2012)Mittal, Moorthy, and Bovik]{MSCN}
	Anish Mittal, Anush~Krishna Moorthy, and Alan~Conrad Bovik.
	\newblock No-Reference Image Quality Assessment in the Spatial Domain.
	\newblock \emph{IEEE Transactions on Image Processing}, 21\penalty0 (12):\penalty0 4695--4708, 2012.
	\newblock \doi{10.1109/TIP.2012.2214050}.
	
	\bibitem[Sun and Li(2025)]{Water-related}
	Zhe Sun and Xuelong Li.
	\newblock Water-related optical imaging: From algorithm to hardware.
	\newblock \emph{Science China(Technological Sciences)}, pages 6--53, 2025.
	
	\bibitem[Wang et~al.(2018)Wang, Dan, and Li]{Multi-perspective}
	Yonghong Wang, Xizuo Dan, Junrui Li, et~al.
	\newblock Multi-perspective digital image correlation method using a single color camera.
	\newblock \emph{Science China(Technological Sciences)}, pages 61--67, 2018.
	
	\bibitem[Yue et~al.(2009)Yue, Li, and Hou]{rocket}
	Chunguo Yue, Jinxian Li, Xiao Hou, et~al.
	\newblock Summarization on variable liquid thrust rocket engines.
	\newblock \emph{Science in China Series E-Technological Sciences}, 52\penalty0 (10):\penalty0 2918--2923, 2009.
	\newblock \doi{10.1007/s11431-009-0185-2}.
	
	\bibitem[Shen et~al.(2022)Shen, Lu, and Pang]{velocimetry}
	Feng Shen, Xinran Lu, Yan Pang, and Zhaomiao Liu.
	\newblock Experimental study on transient flow patterns in simplified saccular intracranial aneurysm models using particle image velocimetry.
	\newblock \emph{Acta Mechanica Sinica}, 38\penalty0 (12):\penalty0 322162, 2022.
	\newblock \doi{10.1007/s10409-022-22162-x}.
	
	\bibitem[Chen et~al.(2025)Chen, Tian, and Jia]{aero-optical}
	Jian Chen, Dapeng Tian, Ping Jia, et~al.
	\newblock Analysis of aero-optical effects within supersonic vehicle imaging fields.
	\newblock \emph{Optics Express}, pages 10262--10278, 2025.
	
	\bibitem[Sun et~al.(2025)Sun, Tang, and Ma]{SUN2025135969}
	Jiuling Sun, Qinglong Tang, Hailong Ma, et~al.
	\newblock Optical diagnostics on the combustion characteristic of ammonia pre-chamber ignition under different thermodynamic boundary conditions.
	\newblock \emph{Energy}, 324:\penalty0 135969, 2025.
	\newblock \doi{10.1016/j.energy.2025.135969}.
	
	\bibitem[Dai et~al.(2025)Dai, Weng, and Liu]{photofragmentation}
	Yan Dai, Wubin Weng, Siyu Liu, et~al.
	\newblock Research progress of laser photofragmentation-fragment detection techniques in combustion diagnostics.
	\newblock \emph{Applied Spectroscopy Reviews}, pages 451--510, 2025.
	
	\bibitem[Han et~al.(2024)Han, He, and Hu]{reconstruction}
	Shihao Han, Yuming He, Yiyu Hu, et~al.
	\newblock Accelerating spectral digital image correlation computation with Taylor series image reconstruction.
	\newblock \emph{Acta Mechanica Sinica}, 40\penalty0 (6):\penalty0 423464, 2024.
	\newblock \doi{10.1007/s10409-024-23464-x}.
	
	\bibitem[Wang et~al.(2025)Wang, Zhang, and Chai]{flows}
	Yan Wang, Jingjing Zhang, Yongfen Chai, et~al.
	\newblock Aerodynamic performance of small wind turbines in sand-laden atmospheric flows.
	\newblock \emph{Acta Mechanica Sinica}, 41\penalty0 (5):\penalty0 324151, 2025.
	\newblock \doi{10.1007/s10409-024-24151-x}.
	
	\bibitem[Zhou(2021)]{review}
	Lixing Zhou.
	\newblock Studies on theory and modeling of droplet and spray combustion in China: a review.
	\newblock \emph{Acta Mechanica Sinica}, 37\penalty0 (7):\penalty0 999--1040, 2021.
	\newblock \doi{10.1007/s10409-021-01124-9}.
	
	\bibitem[Malvar et~al.(2004)Malvar, He, and Cutler]{Malvar2004}
	H.S. Malvar, Li-wei He, and R.~Cutler.
	\newblock High-quality linear interpolation for demosaicing of Bayer-patterned color images.
	\newblock In \emph{2004 IEEE International Conference on Acoustics, Speech, and Signal Processing}, volume~3, pages iii--485, 2004.
	\newblock \doi{10.1109/ICASSP.2004.1326587}.
	
	\bibitem[Guan and Zhao(2025)]{guanpami25}
	Banglei Guan and Ji Zhao.
	\newblock Affine Correspondences between Multi-Camera Systems for Relative Pose Estimation.
	\newblock \emph{IEEE Transactions on Pattern Analysis and Machine Intelligence}, pages 1--18, 2025.
	\newblock \doi{10.1109/TPAMI.2025.3626134}.
	
\end{thebibliography}
\end{document}